\documentclass[10pt,journal,compsoc]{IEEEtran}
\usepackage{graphicx}
\usepackage{amsmath}
\usepackage{amssymb}
\usepackage{multirow}
\usepackage[dvipsnames, svgnames, x11names]{xcolor}
\usepackage{floatrow}
\usepackage{bbding}
\usepackage{pifont}
\usepackage{wasysym}
\usepackage{amssymb}
\usepackage{pifont}
\usepackage{bm}

\usepackage{hyperref}
\hypersetup{hidelinks,
	pdfstartview=Fit,
	breaklinks=true}
\usepackage{url}

\newcommand{\cblue}[1]{{\color{black} #1}}   
\newcommand{\gjtwo}[1]{{\color{black} #1}}    
\newcommand{\gjthree}[1]{{\color{black} #1}}  
\newcommand{\gjfour}[1]{{\color{black} #1}}    
\newcommand{\gjfive}[1]{{\color{black} #1}}    
\newcommand{\gjsix}[1]{{\color{black} #1}}    
\newcommand{\gjseven}[1]{{\color{black} #1}}    
\newcommand{\gjeight}[1]{{\color{black} #1}}    

\newcommand{\gjreone}[1]{{\color{black}#1}}    
\newcommand{\gjretwo}[1]{{\color{black}#1}}    

\newcommand{\gjrelast}[1]{{\color{black}\textit{#1}}}    

\newcommand{\gjnine}[1]{{\color{black} #1}}    
\newcommand{\gjten}[1]{{\color{black} #1}}    

\newcommand{\gjele}[1]{{\color{black} #1}}
\newcommand{\gjtwleve}[1]{{\color{black} #1}}
\newcommand{\gjthirteen}[1]{{\color{black} #1}}

\newcommand{\cmark}{\ding{51}}%
\newcommand{\xmark}{\ding{55}}%

\newcommand{\ie}{\textit{i.e.}, }

\newcommand{\tabincell}[2]{\begin{tabular}{@{}#1@{}}#2\end{tabular}}

\ifCLASSOPTIONcompsoc
  \usepackage[nocompress]{cite}
\else
  \usepackage{cite}
\fi
\ifCLASSINFOpdf
\else
\fi
\begin{document}

{\onecolumn

\noindent \vspace{1cm}

\noindent \textbf{\huge{PSLT: A Light-weight Vision Transformer with \\ \\ Ladder Self-Attention and Progressive Shift}}

\vspace{2cm}

\noindent {\LARGE{Gaojie Wu, Wei-Shi Zheng*, Yutong Lu, Qi Tian}}
\\
\\
*Corresponding author: Wei-Shi Zheng.
\\
\\
Project page with code: \url{https://isee-ai.cn/wugaojie/PSLT.html}

\vspace{1cm}

\noindent {\LARGE{Submission date: 08-Jul-2022 to IEEE Transaction on Pattern Analysis and Machine Intelligence
}}

\vspace{1cm}

\noindent For reference of this work, please cite:

\vspace{1cm}
\noindent Gaojie Wu, Wei-Shi Zheng, Yutong Lu and Qi Tian.
``PSLT: A Light-weight Vision Transformer with Ladder Self-Attention and Progressive Shift''. \emph{IEEE Transaction on Pattern Analysis and Machine Intelligence,} 2023.

\vspace{1cm}

\noindent Bib:\\
\noindent @article\{wu2023pslt,\\
\ \ \  title     = \{PSLT: A Light-weight Vision Transformer with Ladder Self-Attention and Progressive Shift\}, \\
 \ \ \   author    = \{Gaojie Wu, Wei-Shi Zheng, Yutong Lu and Qi Tian\},\\
\ \ \  journal   = \{\{IEEE\} Transaction on Pattern Analysis and Machine Intelligence\},\\
\ \ \  year      = \{2023\}\\
\}
}

\twocolumn

%
\title{PSLT: A Light-weight Vision Transformer with Ladder Self-Attention and Progressive Shift}
%
%
%
%

\author{Gaojie Wu,
        Wei-Shi Zheng*, \thanks{* Corresponding author}
        Yutong Lu
        and Qi Tian
\IEEEcompsocitemizethanks{\IEEEcompsocthanksitem
Gaojie Wu and Yutong Lu are with the School of Computer Science and Engineering, Sun Yat-sen University, Guangzhou 510275, China. E-mail:
wugj7@mail2.sysu.edu.cn, yutong.lu@nscc-gz.cn.
\IEEEcompsocthanksitem
Wei-Shi Zheng is with the School of Data and Computer Science, Sun Yat-sen University, Guangzhou 510275, China, with Peng Cheng Laboratory, Shenzhen 518005, China, and also with the Key Laboratory of Machine Intelligence and Advanced Computing (Sun Yat-sen University), Ministry of Education, China. E-mail: wszheng@ieee.org /zhwshi@mail.sysu.edu.cn.
\IEEEcompsocthanksitem
Qi Tian is with the Cloud $\&$ AI BU, Huawei, China (tian.qi1@huawei.com).
}
}

%
%

\markboth{SUBMISSION TO IEEE TRANSACTIONS ON PATTERN ANALYSIS AND MACHINE INTELLIGENCE}%
{Shell \MakeLowercase{\textit{et al.}}: Bare Demo of IEEEtran.cls for Computer Society Journals}
%



\IEEEtitleabstractindextext{%
\begin{abstract}

Vision Transformer (ViT) has shown great potential for various visual tasks due to its ability to model long-range dependency. However, ViT requires a large amount of computing resource to compute the global self-attention. In this work, we propose a ladder self-attention block with multiple branches and a progressive shift mechanism to develop a light-weight transformer backbone that requires less computing resources (e.g. a relatively small number of parameters and FLOPs), termed Progressive Shift Ladder Transformer (PSLT). \gjseven{First,} the ladder self-attention block reduces the computational cost by modelling local self-attention in each branch. \gjseven{In the meanwhile, the progressive shift mechanism is proposed to enlarge the receptive field in the ladder self-attention block by modelling diverse local self-attention for each branch and interacting among these branches.}
\gjseven{Second, the input feature of the ladder self-attention block is split equally along the channel dimension for each branch, which considerably reduces the \gjthirteen{computational cost} in the ladder self-attention block (with nearly $\frac{1}{3}$ the amount of parameters and FLOPs), and the outputs of these branches are then collaborated by a pixel-adaptive fusion}. 
\gjseven{Therefore, the ladder self-attention block with a relatively small number of parameters and FLOPs is capable of modelling long-range interactions.} 
Based on the ladder self-attention block,
\cblue{PSLT performs well on several vision tasks, including image classification, objection detection and person re-identification. On the ImageNet-1k dataset, PSLT achieves a top-1 accuracy of 79.9\% with 9.2M parameters \gjnine{and 1.9G FLOPs}, which is comparable to several existing models with more than 20M parameters \gjnine{and 4G FLOPs}.} Code is available at 
\url{https://isee-ai.cn/wugaojie/PSLT.html}.
\end{abstract}

}

\maketitle

\IEEEdisplaynontitleabstractindextext

%
\IEEEpeerreviewmaketitle

\IEEEraisesectionheading{\section{Introduction}\label{sec:introduction}}

Convolutional neural networks (CNNs) have shown considerable promise as general-purpose backbones for computer vision tasks. Since AlexNet \cite{krizhevsky2012imagenet} was proposed for the ImageNet image classification challenge \cite{Russakovsky2015ImageNet}, CNN architectures have become increasingly powerful, with more careful designs \cite{he2016deep, gao2019res2net, xie2017aggregated}, deeper connections \cite{simonyan2014very, huang2017densely} and wider dimensions \cite{zagoruyko2016wide}. Thus, CNNs appear to be indispensable for various computer vision tasks. CNNs are successful due to the inductive bias implied in convolutional computations, and this inductive bias ensures that CNNs are generalizable \gjseven{as investigated in \cite{dai2021coatnet, xiao2021early}}.

Another prevalent architecture that has shown extraordinary achievements in computer vision tasks is the vision transformer. In the transformer, which was first proposed for sequence modelling and transduction tasks in natural language processing \cite{vaswani2017attention, devlin2018bert}, features can interact globally by modelling attention of long-range dependency.
\gjseven{Vision Transformer (ViT) models global interaction by computing attention among the feature map according to the projected query, key and value of each pixel.} ViT has shown these achievements due to the considerable amount of computing resource in the model. However, models with large number of parameters are difficult to deploy in edge-computing devices with limited memory storage and computing resources, such as FPGAs. \gjnine{And models with large number of FLOPs always require much time for inference, because FLOPs measures the number of float-point operations for inference.}

In this work, we aim to develop a light-weight vision transformer backbone with a relatively small number of parameters \gjnine{and FLOPs}. Existing methods \cite{guo2021cmt, wang2021pyramid, liu2021swin, wang2021crossformer} have focused on reducing the number of float-point operations by evolving the form of computing self-attention; 
\gjfive{however, the receptive field in the window-based self-attention block is restricted in most of these methods \cite{liu2021swin, wang2021crossformer}, and only pixels divided in the same windows can interact with each other in the window-based self-attention block. 
Therefore, the interactions between pixels in different windows cannot be modelled in one block.}


Thus, we propose a light-weight ladder self-attention block with multiple branches and a progressive shift mechanism is introduced to enlarge the receptive field explicitly. \gjnine{The expansion of the receptive field for the ladder self-attention block is accomplished according to the following strategy. First, diverse local self-attentions are modelled by steering pixels at the same spatial position to model interactions with pixels in diverse windows for different branches. Second, the progressive shift mechanism transmits the output features of the current branch to the subsequent branch. \gjten{The self-attention in the subsequent branch is computed with the participation of the output features in the current branch,} allowing interactions among features in different windows in the two branches.} As a result, the ladder self-attention block with the progressive shift mechanism can model long-range interactions among pixels divided in different windows. In addition, in the ladder self-attention block, each branch only takes an equal proportion of input channels of the block, which considerably reduces the number of parameters and FLOPs. And all the channels in these branches are aggregated in the proposed pixel-adaptive fusion module to generate the output of the ladder self-attention block. Based on the above designs, we developed a light-weight general-purpose backbone with a relatively small number of parameters.

The overall framework of our model is shown in Figure \ref{fig:structure}. \gjnine{According to the conclusion of \cite{dai2021coatnet, xiao2021early} that the convolution in the early stages helps the model learn better and faster. PSLT adopts light-weight convolutional blocks in the early stages and the ladder self-attention blocks in the latter stages. Our PSLT has several important characteristics:}
\begin{enumerate}
    \item \gjthree{\gjfive{PSLT uses light-weight ladder self-attention blocks, which greatly reduce the number of trainable parameters and FLOPs.} The ladder self-attention block first divides the input feature map into several equal proportions along the channel axis. Then, each part of the feature map is sent to an individual branch to compute the self-attention similarity.}

    
    \item \gjseven{The ladder-self attention block is designed to acquire large receptive field, requiring relatively small number of computing resources.} \gjnine{PSLT models local attention on each branch for computing efficiency;
    and more importantly, without introducing extra parameters, PSLT adopts the progressive shift mechanism to model interactions among pixels in different windows to enlarge the receptive field of each ladder self-attention block. The progressive shift mechanism forms a block shaped like a ladder.} 
    
\end{enumerate}



Our proposed PSLT achieves excellent performance on visual tasks such as image classification, object detection and person re-identification with a relatively small number of trainable parameters. With less than 10 million parameters and \gjnine{2G FLOPs}, PSLT achieves a top-1 accuracy of 79.9\% on ImageNet-1k image classification at input resolution $224\times 224$  \gjnine{without pretraining on extra large dataset}.

\gjele{We notice that recently there exist light-weight transformers \cite{pan2022edgevits, maaz2022edgenext, mehta2021mobilevit} that combine local interaction (convolution) and global self-attention in one block. In this work, we provide another perspective, and our PSLT is capable of modelling long-range interaction by effectively incorporating diverse local self-attentions and \gjtwo{evolving} the self-attention of the latter branches in the ladder self-attention block with a relatively small number of parameters and FLOPs.}

In summary, our contributions are as follows:
\begin{itemize}

\item \gjthree{We propose a light-weight ladder self-attention block with multiple branches and considerably less parameters. A pixel-adaptive fusion module is developed to aggregate features from multiple branches with adaptive weights along both the spatial and channel dimensions.

\item \gjsix{We propose a progressive shift mechanism in the ladder self-attention block to enlarge the receptive field. The progressive mechanism not only allows modelling interaction among pixels in different windows for long-range dependencies, but also reduces the number of parameters necessary to obtain the value for computing self-attention in the following branch.} }

\item We develop a general-purpose backbone (PSLT) with a relatively small number of parameters. PSLT is constructed with light-weight convolutional blocks and the ladder self-attention blocks, improving its generalization ability.

\end{itemize}

\section{Related Work} \label{section:relatedWork}

\gjtwo{A comparison of various models is shown in Table \ref{tab:comparison}.
Our PSLT was manually developed and has a relatively small number of parameters, and PSLT does not utilize a search phase before training. In the following section, we detail some related works.
 }

\subsection{Convolutional Neural Networks}
Since the development of AlexNet \cite{krizhevsky2012imagenet}, CNNs have become the most popular general-purpose backbones for various computer vision tasks. More effective and deeper convolutional neural networks have been proposed to further improve the model capacity for feature representation, such as VGG \cite{simonyan2014very}, ResNet \cite{he2016deep}, DenseNet \cite{huang2017densely} and ResNext \cite{xie2017aggregated}. However, neural networks with large scales are difficult to deploy in edge devices, which have limited memory resources. Thus, various works have focused on designing light-weight neural networks with less trainable parameters.

\vspace{0.1cm}

\noindent \textbf{- Light-weight CNN.} To decrease the number of trainable parameters, MobileNets \cite{howard2017mobilenets, sandler2018mobilenetv2, howard2019searching} substitute the standard convolution operation with a more efficient combination of depthwise and pointwise convolution. ShuffleNet \cite{zhang2018shufflenet} uses group convolution and channel shuffle to further simplify the model. The manual design of neural networks is time consuming, and automatically designing convolutional neural architectures has shown great potential for developing high-performance neural networks \cite{zoph2017neural, guo2020single, li2021bossnas}. EfficientNet \cite{tan2019efficientnet} investigates the model width, depth and resolution with a neural architecture search \cite{zoph2017neural, guo2020single, liu2019darts}.

Although CNNs can effectively model local interactions, adaptation to input data is absent in standard convolution. \gjreone{Usually, the adaptive methods are capable of improving model capacity by the dynamic weights or receptive fields with respective to different input images.} RedNet \cite{li2021involution} produces dynamic weights for distinct input data. Deformable convolution \cite{dai2017deformable} produces an adaptive location for the convolution kernels. Our proposed backbone (PSLT) 
inherits the virtue of convolution that implies inductive bias when modelling local interactions, and PSLT takes advantage of self-attention \cite{vaswani2017attention} to model long-range dependency and adapt to input data \gjretwo{with dynamic weights}.

\begin{table}[ht]
    \centering
    \caption{Comparison of various general-purpose backbones. ``T'' indicates that the model adopts the transformer architecture. ``Adaptive'' indicates that the model can adapt to the input image. ``Light'' denotes that the model has a relatively small number of parameters and FLOPs. ``\gjretwo{NAS}'' denotes that the model needs another search process before training from scratch.}
    \setlength{\tabcolsep}{0.7mm}
    {\begin{tabular}{c|c|c|c|c}
        \hline
        Model & Architecture & Adaptive & Light & \tabincell{c}{\gjretwo{NAS}} \\
        \hline
        VGG \cite{simonyan2014very} & CNN & \xmark & \xmark & \xmark \\
        ResNet \cite{he2016deep} & CNN & \xmark & \xmark & \xmark \\
        DenseNet \cite{huang2017densely} & CNN & \xmark & \xmark & \xmark \\
        MobileNet \cite{howard2017mobilenets} & CNN & \xmark & \cmark & \xmark \\
        ShuffleNet \cite{zhang2018shufflenet} & CNN & \xmark & \cmark & \xmark \\
        EfficientNet & CNN & \cmark & \xmark & \cmark \\
        RedNet \cite{li2021involution} & CNN & \cmark & \xmark & \xmark \\
        \hline
        ViT \cite{dosovitskiy2020image} & T & \cmark & \xmark & \xmark \\
        PS-ViT \cite{yue2021vision} & T & \cmark & \xmark & \xmark \\
        Swin \cite{liu2021swin} & T & \cmark & \xmark & \xmark \\
        VOLO \cite{yuan2021volo} & T & \cmark & \xmark & \xmark \\
        PVT \cite{wang2021pyramid} & T & \cmark & \xmark & \xmark \\
        MobileViT \cite{mehta2021mobilevit} & T & \cmark & \cmark & \xmark \\
        EfficientFormer \cite{li2022efficientformer} & T & \cmark & \cmark & \cmark \\
        \hline
        CoATNet & CNN+T & \cmark & \xmark & \xmark \\
        CvT \cite{wu2021cvt} & CNN+T & \cmark & \xmark & \xmark \\
        CMT \cite{guo2021cmt} & CNN+T & \cmark & \xmark & \xmark \\
        Conformer \cite{peng2021conformer} & CNN+T & \cmark & \xmark & \xmark \\
        BossNet \cite{li2021bossnas} & CNN+T & \cmark & \xmark & \cmark \\
        LeViT \cite{graham2021levit} & CNN+T & \cmark & \cmark & \xmark \\
        Mobile-Former \cite{chen2021mobile} & CNN+T & \cmark & \cmark & \xmark \\
        \hline
        \textbf{PSLT (Ours)} & CNN+T & \cmark & \cmark & \xmark \\
        \hline
    \end{tabular}}
    \label{tab:comparison}
\end{table}

\subsection{Vision Transformer}
Transformers \cite{vaswani2017attention, devlin2018bert} are widely used in natural language processing, and the transformer architecture for computer vision tasks has evolved since ViT \cite{dosovitskiy2020image} was proposed for image classification. ViT has one-stage structure that produces features with only one scale. PiT \cite{heo2021rethinking} adopts depthwise convolution \cite{howard2017mobilenets} to decrease the spatial dimension. CrossViT \cite{chen2021crossvit} models multi-scale features via small and large patch embedding sizes. Recent transformer architectures \cite{liu2021swin, wang2021pyramid, graham2021levit, guo2021cmt, fan2021multiscale} have adopted hierarchical structures like popular CNNs to yield multi-scale features through patch merging to progressively reduce the number of patches. \gjthirteen{To mitigate the issue that the tokenization in ViT \cite{yuan2021tokens} may destruct the object structure, PS-ViT \cite{yue2021vision} locates discriminative regions by iteratively sampling tokens.}

Since self-attention requires a large amount of computing resources to model long-range interactions, recent works \cite{liu2021swin, wang2021crossformer, yu2021glance} have approximated global interaction by modelling local and long-range dependency. SwinTransformer \cite{liu2021swin} divides the feature map into multiple windows, self-attention is only conducted for interactions among pixels in the same window, and the shift operation is proposed for long-range interactions. VOLO \cite{yuan2021volo} generates the outlook attention matrix for a local window. DAT \cite{xia2022vision} generates deformable attention similar to deformable convolution \cite{dai2017deformable}. Twins \cite{chu2021twins} first reduces the size of the feature map and then conducts self-attention and upsamples the feature map to its original size. PVT \cite{wang2021pyramid}, ResT \cite{zhang2021rest} and CMT \cite{guo2021cmt} maintain the size of the query while reducing the size of the key and value to process the self-attention operation. CSWin \cite{dong2022cswin} adopts the cross-shaped window self-attention mechanism for computing efficiency.

\vspace{0.3cm}

\begin{table}[t]
    \centering
    \caption{\gjele{Comparison of light-weight vision transformers. ``T'' indicates that the model adopts the transformer architecture. ``Local'' indicates the modelling of local interaction. ``Global'' indicates the modelling of long-range interaction. ``GSA'' denotes that the model adopts global self-attention. ``LSA'' denotes that the model adopts local self-attention}.}
    {\begin{tabular}{c|c|c|c}
        \hline
        Model & Architecture & Local & Global \\
        \hline
        Mobile-Former \cite{chen2021mobile} & CNN+T & CNN & GSA \\
        MobileViT \cite{mehta2021mobilevit} & CNN+T & CNN & GSA \\
        EdgeViT \cite{pan2022edgevits} & CNN+T & CNN & GSA \\
        EdgeNext \cite{maaz2022edgenext} & CNN+T & CNN & GSA \\
        \hline
        \textbf{PSLT (Ours)} & CNN+T & LSA & LSA \\
        \hline
    \end{tabular}}
    \label{tab:comparison-light}
\end{table}

\noindent \textbf{- CNN + Transformer}
The combination of convolution and self-attention has shown great potential in computer vision tasks. CoATNet \cite{dai2021coatnet} investigates the best method for allocating the convolutional blocks and transformer blocks in different stages to achieve the optimal model generalization ability and capacity. BossNet \cite{li2021bossnas} uses a neural architecture search method \cite{zoph2017neural} to automatically determine the best method to combine the convolutional block and transformer block. LeViT \cite{graham2021levit} proposes a hybrid neural network for fast inference image classification. LeViT utilizes convolution with a kernel size of 3 to halve the feature size four times and downsamples during the self-attention operation to reduce the number of float-point operations. CvT \cite{wu2021cvt} introduces depthwise and pointwise convolution to replace the fully connected layer, yielding the query, key and value in the multi-head self-attention mechanism. Conformer \cite{peng2021conformer} uses a dual network structure with convolution and self-attention for enhanced representation learning. CMT \cite{guo2021cmt} also introduces depthwise and pointwise convolution in multi-head self-attention mechanism, and an inverted residual FFN is used in place of a standard FFN for improvement. \gjten{MPViT \cite{lee2021mpvit} combines the convolution and multi-scale self-attention for multi-scale representation.}

\noindent \textbf{- Light-weight vision transformers.} The achievements of vision transformers rely on the large number of computing resource (parameters and FLOPs). Mobile-Former \cite{chen2021mobile} leverages the advantages of MobileNet for local processing and the advantages of the transformer for global interaction. \gjten{EdgeViT \cite{pan2022edgevits} and EdgeNeXt \cite{maaz2022edgenext} also combine local interaction (convolution) and global self-attention.} MobileViT \cite{mehta2021mobilevit} is also a light-weight general-purpose vision transformer for mobile devices. MobileViT leverages a multi-scale sampler for efficient training. LVT \cite{yang2022lite} also develops enhanced self-attention for computing efficiency.

\gjele{In this work, we propose a light-weight ladder self-attention block requiring a relatively small number of parameters and FLOPs. Instead of combining convolution and global self-attention for local and global interactions in a block, our ladder self-attention block enlarges the receptive field by modelling and interacting the diverse local self-attentions, as shown in Table \ref{tab:comparison-light}. And modelling local self-attention is more computationally efficient than modelling global self-attention. Then we develop a light-weight transformer backbone with the ladder self-attention block in the last two stages and light-weight convolutional blocks in the first two stages,} and the allocation of self-attention blocks and convolutional blocks is proved with better performance in CoAtNet \cite{dai2021coatnet}.


\section{Progressive Shift Ladder Transformer (PSLT)}
\label{sec:mehod}

\begin{figure*}[t]
\begin{center}
  \includegraphics[width=\linewidth]{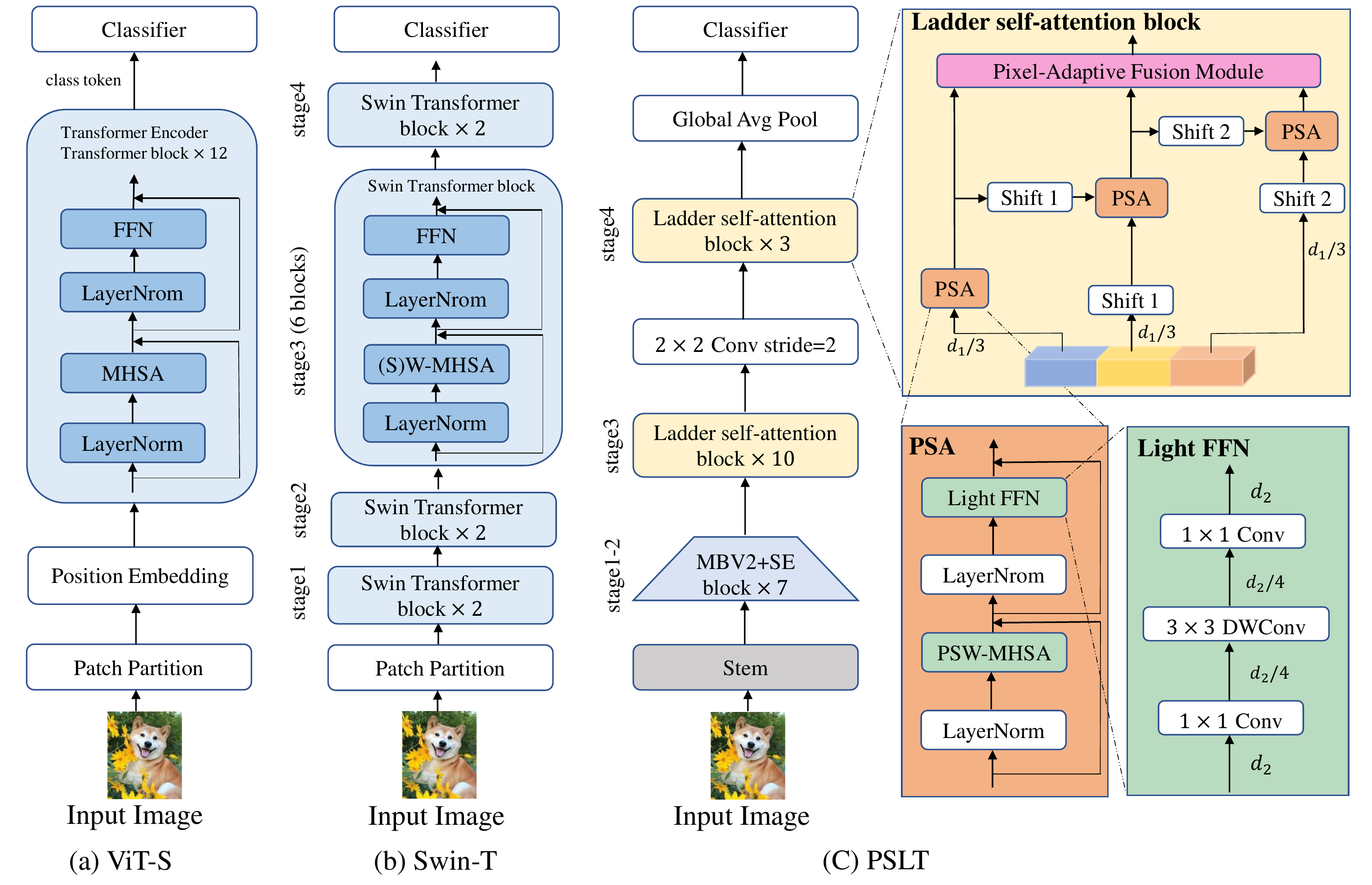}
\end{center}
\caption{Illustration of prevalent transformer architectures for image classification. (a) ViT \cite{dosovitskiy2020image} with only one stage. (b) Swin Transformer \cite{liu2021swin}, which uses window-based multi-head self-attention and multi-stage blocks. (c) \gjtwo{Our proposed PSLT adopts light-weight ladder self-attention (SA) blocks in the final two stages to model long-range dependency of pixels divided in different windows in the same block, and the light-weight convolutional blocks (MobileNetV2 \cite{sandler2018mobilenetv2} block with a squeeze-and-excitation (SE) block \cite{hu2018squeeze}) are adopted in the first two stages.} The details of PSLT are described in Section \ref{sec:mehod}, and the structures of the PSW-MHSA and pixel-adaptive fusion module are shown in Figure \ref{fig:multi-branch-SA}. ``Shift 1'' and ``Shift 2'' denotes shift operations with different directions. ``$d_1$'' and ``$d_2$'' denote the number of channels, and $d_1=3d_2$.}
\label{fig:structure}
\end{figure*}

\subsection{Overall Architecture}
\label{sec:architecture}

In this work, we aim to develop a light-weight transformer backbone and expand the receptive field of the basic window-based self-attention block. We propose a ladder self-attention block with multiple branches. Each branch takes an equal proportion of input features and adopts the window-based self-attention, considerably reducing the number of parameters and float-point operations in the ladder self-attention block. 

A progressive shift mechanism is formed to enlarge the receptive field of the ladder self-attention block with the following strategy. Each branch shifts the obtained features in different directions and divides the shifted features into multiple windows. In this manner, the output features of each branch can be aggregated from diverse windows.
\gjnine{Furthermore, the latter branch takes the output feature of the previous branch as input for computing self-attention, allowing pixels in different windows of various branches to interact with one another.} \gjtwleve{With the progressive shift mechanism, the light-weight ladder self-attention block is capable of modelling long-range interactions.} Because the output feature of each branch is integrated with pixels in different spatial windows, a pixel-adaptive fusion module is developed to effectively integrate the output features of all the branches with adaptive weights along both the spatial and channel dimensions.

Based on the above considerations, we finally develop a light-weight general-purpose backbone with a relatively small number of trainable parameters based on the proposed ladder self-attention block, termed Progressive Shift Ladder Transformer (PSLT). An overview of PSLT is shown in Figure \ref{fig:structure}. PSLT leverages the advantages of convolution for local interaction and self-attention for long-range interaction. An input image is passed through a stem convolution layer and four stages consisting of convolutional blocks or the proposed ladder self-attention blocks. The output feature of the final stage is sent to the global average pooling layer and classifier.


To ensure that PSLT is applicable to various computer vision tasks, we adopt a four-stage architecture following the Swin Transformer \cite{liu2021swin} to yield the hierarchical features. 
To improve the model generalization ability and capacity, PSLT adopts light-weight convolutional blocks in MobileNetV2 \cite{sandler2018mobilenetv2} with a squeeze-and-excitation (SE) block \cite{hu2018squeeze} in the first two stages and ladder self-attention in the final two stages. In the final two stages, the proposed light-weight ladder self-attention blocks are applied to model long-range interactions.
Note that the input feature is downsampled at the beginning of each stage, and these stages jointly yield multi-scale features similar to prevalent convolutional architectures. With these hierarchical representations, PSLT can be easily applied as the backbone model in existing frameworks for various computer vision tasks. For image classification, the output feature of the final stage, with abstract semantics representing the global information, is transmitted to the global average pooling layer and classifier.

In addition, in contrast to splitting the input RGB image into non-overlapping patches (Patch Partition) for the initial representation, as is performed in most commonly used vision transformer architectures, PSLT applies the popular stem with three convolution layers for feature extraction, as splitting the image into non overlapping patches may divide the same part of an object into different patches. 
The $3\times3$ convolution in the first process is capable of modelling local interactions with the implied inductive bias.

\gjtwo{Different from existing light-weight transformer \cite{chen2021mobile}, which greatly reduces the number of parameters in backbones and introduces multiple parameters in the classification head, PSLT has less parameters in the classification head. Our experiments show that PSLT uses only 6\% (0.57 M of 9.2 M) of the parameters in the classification head, which is considerably lower than Mobile-Former-294M \cite{chen2021mobile} using 40\% (4.6 M of 11.4 M) of the parameters in the classification head of the two fully connected layers.}

\begin{figure}[t]
\begin{center}
  \includegraphics[width=\linewidth]{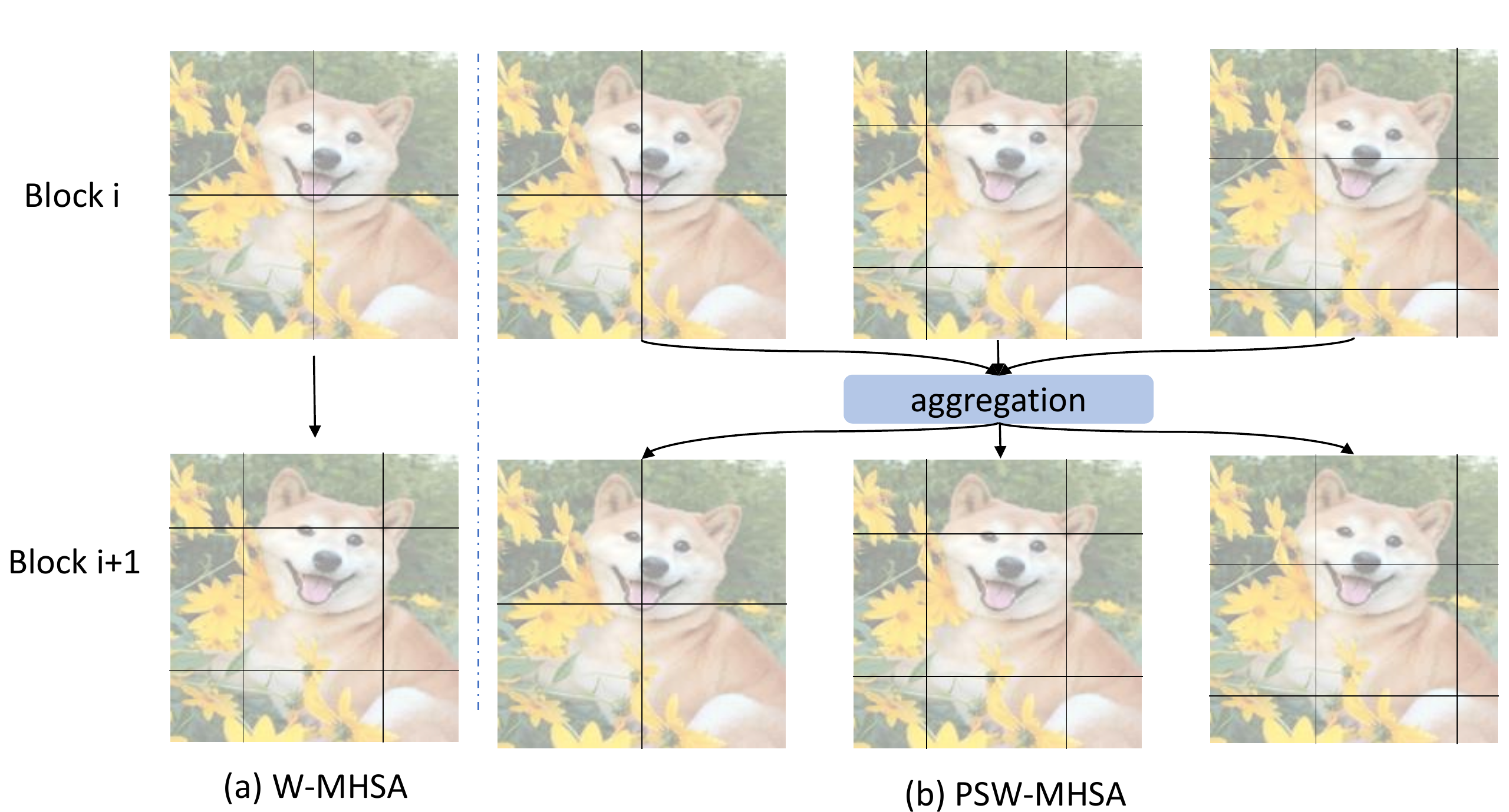}
\end{center}
\caption{\gjfour{Illustration of the window partition for pixel interactions in each block. (a) The original W-MHSA generates windows according to only one strategy to model local attention in each block. (b) The PSW-MHSA produces different windows with multiple strategies and fuses these features to aggregate information from diverse spatial windows.}}
\label{fig:window}
\end{figure}

\begin{figure*}[t]
\begin{center}
  \includegraphics[width=\linewidth]{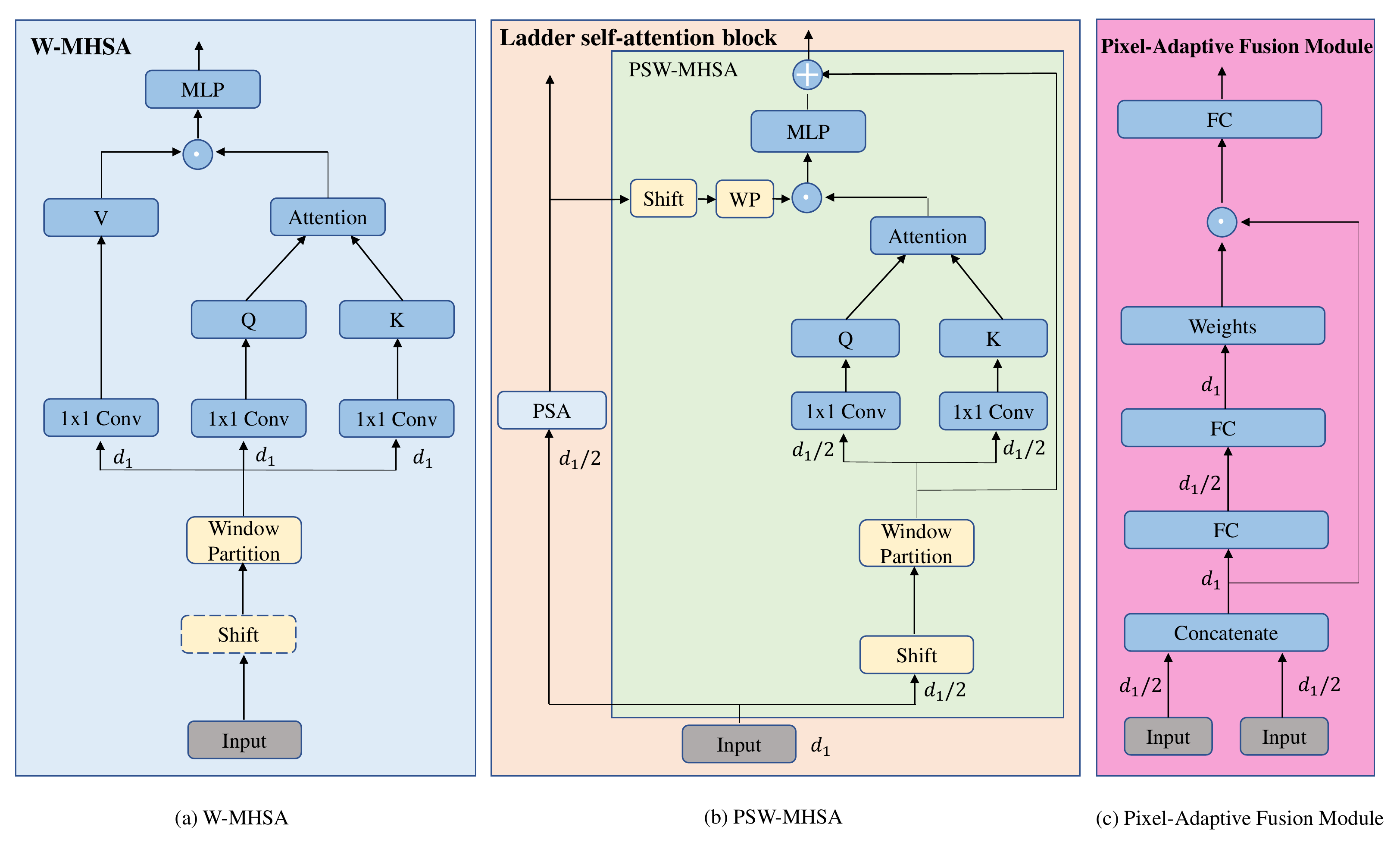}
\end{center}
\caption{Example of detailed blocks in PSLT. (a) Window-based multi-head self-attention in the Swin Transformer \cite{liu2021swin}. (b) The ladder self-attention block with two branches for example. \gjeight{The progressive shift mechanism transmits output feature of the previous branch to the subsequent branch, which further enlarges the receptive field.} Only the PSW-MHSA is illustrated in detail; \cblue{LayerNrom and Light FFN are shown in Figure \ref{fig:structure} and omitted here for simplicity}. ``WP'' indicates the window partition. (c) The pixel-adaptive fusion module in the ladder self-attention block. ``Weights'' denotes the adaptive weights along the channel and spatial dimensions, which is the output of the second fully connected layer. The details of the PSW-MHSA are described in Section \ref{sec:progressive}.}
\label{fig:multi-branch-SA}
\end{figure*}

\subsection{Ladder Self-Attention Block}
\label{sec:multi-branch}
\gjthree{To model interaction among pixels in different windows, we propose a ladder self-attention block with multiple branches and model long-range interaction among pixels in different branches through the progressive shift mechanism. Since the feature maps are integrated in diverse spatial windows in each branch of the ladder self-attention block, a pixel-adaptive fusion module is developed for effective feature fusion among these branches. \gjfour{
The difference of processing feature maps in these branches allows the proposed block to obtain diverse information, and the proposed pixel-adaptive fusion module is capable of efficiently aggregating the diverse information.} In the following section, we describe the details of the ladder self-attention block.}



\subsubsection{Progressive Shift for the Ladder Self-Attention Block}
\label{sec:progressive}
The ladder self-attention block divides the input feature map into several equal proportions along the channel dimension and sends these features to multiple branches. Take the ladder self-attention block in Figure \ref{fig:structure} for example; the input feature map with $d$ channels is divided into three parts, each with $\frac{d}{3}$ channels for each branch. PSLT adopts window-based self-attention to only compute similarity among pixels in the same windows, thus ensuring computing efficiency. 
\gjeight{Different from SwinTransformer \cite{liu2021swin} that stacks even number of window-based self-attention blocks to enlarge the receptive field with shift operation, our PSLT adopts the progressive shift mechanism to model diverse local interactions in each branch and to interact information among these branches, which explicitly enlarges the receptive field of each ladder self-attention block. In this manner, PSLT acquires large receptive field with a relatively small number of parameters and FLOPs.}

For long-range interaction, PSLT proposes a progressive shift mechanism for the ladder self-attention block. First, \gjtwleve{the input features in the first branch are processed by the PSA with the W-MHSA without the shift operation, as shown in Figure \ref{fig:multi-branch-SA} (a).} And the features of the other two branches are shifted differently to \gjsix{model diverse local interactions, namely pixels from two windows in one branch may be divided in the same window in the other branches.} The shift direction or stride of the two branches is different from each other to model long-range interactions. \gjfour{An illustration of modelling interaction in different windows is shown in Figure \ref{fig:window}.} Second, the progressive shift manner transmits the output feature of the current branch to the subsequent branch. As shown in Figure \ref{fig:multi-branch-SA} (b), the PSW-MHSA takes both the output feature of the previous branch and the current input feature as input for self-attention computation to model interactions among the branches.

More specifically, the PSW-MHSA applies the same shift and window partition operations to both the input feature of the current branch and the output feature of the previous branch. A $1\times1$ convolution is applied to the input feature to yield the query and key to compute the similarity among the feature points. Instead of yielding the value in the same manner, the PSW-MHSA takes the output of the previous branch as the value directly for the self-attention computation. 
The PSW-MHSA progressively delivers the output features of the branches, allowing it to model long-range dependency because after the local interactions are modelled, the pixels constrained in the divided windows of the current branch can interact with pixels outside the window in the following branches. 

Instead of consecutive connection of local self-attention and shifted local self-attention for long-range interaction \cite{liu2021swin}, our PSW-MHSA enables building backbone with arbitrary number of ladder self-attention blocks instead of even number of blocks. \gjthree{In this way, the ladder self-attention block can model long-range interactions among pixels in different windows with the help of multiple branches, and the training cost (number of trainable parameters) is decreased .}

Formally, the ladder self-attention blocks are computed as
\begin{equation}
    \begin{aligned}
    &\hat{O}_t = \textnormal{PSW-MHSA}(I_t, O_{t-1}) \\
    \textnormal{PSW-MHSA}&(I_t, O_{t-1})=\textnormal{softmax}(\frac{Q_{I_t}K_{I_t}}{\sqrt{d}})O_{t-1} + I_t \\
    &O_t=\textnormal{LFFN}(\textnormal{LN}(\hat{O}_t)) \\
    &\Bar{O} = \textnormal{PAFM}(O_t), t=0,1,2,....
    \end{aligned}
\end{equation}
The output ($\hat{O}_t$) of the PSW-MHSA in the $t$-th branch is computed according to the input feature of the $t$-th branch ($I_t$) and the output feature of the $(t-1)$-th branch ($O_{t-1}$). In the PSW-MHSA, the query ($Q_{I_t}$) and the key ($K_{I_t}$) are projected from $I_t$, and $O_{t-1}$ is directly applied as the value. The first branch is computed as $O_0=\textnormal{MHSA}(I_0)$, as shown in Figure \ref{fig:multi-branch-SA} (a). Then, the Light FFN (LFFN) and layer norm (LN) are applied to produce the output of the $t$-th branch ($O_t$). Finally, a pixel-adaptive fusion module (PAFM) is developed to generate the output of the ladder self-attention block ($\Bar{O}$) according to the outputs of all branches.

\vspace{0.1cm}

\noindent \textbf{- Complexity analysis.} Our intention is to build a light-weight backbone, and the PSW-MHSA is proposed for modelling long-range dependency. Here, we roughly compare the number of trainable parameters and computation complexity of the W-MHSA and PSW-MHSA. Taking the input of $h\times w\times d_1$ as an example, each window has $M\times M$ non-overlapping patches, and $B$ branches are contained in the PSW-MHSA. Then, the number of trainable parameters in the W-MHSA and PSW-MHSA are:
\begin{equation}
    \begin{aligned}
        \phi (\textnormal{W-MHSA}) &= 4d_1^2 \\
        \phi (\textnormal{PSW-MHSA}) &= \frac{3d_1^2}{B} + \frac{d_1^2}{B^2} .
    \end{aligned}
\end{equation}
The computational complexity of the W-MHSA and PSW-MHSA are:
\begin{equation}
    \begin{aligned}
        \Omega (\textnormal{W-MHSA}) &= 4hwd_1^2 + 2M^2hwd_1 \\
        \Omega (\textnormal{PSW-MHSA}) &= \frac{3hwd_1^2}{B} + \frac{hwd_1^2}{B^2} + 2M^2hwd_1 .
    \end{aligned}
\end{equation}
As can be seen, when $B=1$, the PSW-MHSA has 
the same number of trainable parameters and complexity as the W-MHSA. 
When $B>1$, the number of trainable parameters and the computation complexity are both considerably reduced ($B$ is set to 3 by default).

\subsubsection{Light FFN}
As presented in Figure \ref{fig:structure} (b), the output of the MHSA is processed by the FFN. To further decrease the number of trainable parameters and float-point operations in PSLT, a Light FFN is proposed to replace the original FFN, as shown in Figure \ref{fig:structure} (c). In contrast to the original FFN, which uses a fully connected layer with $d$ channels for both the input and output features, the Light FNN first projects the input with $d_2$ channels to a narrower feature with $\frac{d_2}{4}$ channels. Then, a depthwise convolution is applied to model local interactions, and a pointwise convolution is adopted to restore the channels. 

\vspace{0.1cm}

\noindent \textbf{- Complexity analysis.} Taking the input of an FFN layer of size $h\times w \times d_2$ as an example, the number of trainable parameters of the FFN and LFFN are:
\begin{equation}
    \begin{aligned}
        \phi (\textnormal{FFN}) &= d^2 \\
        \phi (\textnormal{LFFN}) &= \frac{d_2^2}{2} + \frac{9d_2}{4} .
    \end{aligned}
\end{equation}
The computational complexity of the FFN and LFFN are:
\begin{equation}
    \begin{aligned}
        \Omega (\textnormal{FFN}) &= hwd_2^2 \\
        \Omega (\textnormal{LFFN}) &= \frac{hwd_2^2}{2} + \frac{9hwd_2}{16} .
    \end{aligned}
\end{equation}
In general, the number of input feature channels $d_2$ is substantially larger than $\frac{9}{2}$; thus, the LFFN is more computationally economical than the FFN, as it has less trainable parameters and float-point operations.

\subsubsection{Pixel-Adaptive Fusion Module}
\label{sec:PAFM}
PSLT divides the input feature map equally along the channel dimension for the multiple branches to decrease the number of parameters and computation complexity. To model long-range dependency, the progressive shift mechanism in the ladder self-attention block divides the same pixel into different windows in each branch. Because the output features of the branches are produced by integrating feature information in different windows, PSLT proposes a pixel-adaptive fusion module to efficiently aggregate multi-branch features with adaptive weights along the spatial and channel dimensions.

More specifically, the output features of all the branches are concatenated and sent to two fully connected layers to yield the weights for each pixel. The weights indicate the importance of the features from all the branches for each pixel in the feature map. To ensure that the number of trainable parameters is not increased, the first fully connected layer halves the number of channels. The features are then multiplied by the weights and then sent to a fully connected layer for feature fusion along the channel dimension. 

\gjtwo{Different from the conventional squeeze-and-excitation block \cite{hu2018squeeze} measuring only the importance of the channels, we introduce the pixel-adaptive fusion module to integrate the output features of all branches \gjseven{with adaptive weights} in both the spatial and channel dimensions for two reasons. First, because the pixels in each branch are divided into different windows, the pixels at various spatial locations contain distinct information. \gjsix{Second, along the channel dimension, the input feature map is split equally for each branch, so the pixel information at the same spatial location is different among the branches.} The experimental results show that our proposed pixel-adaptive fusion module is more suitable for feature fusion than the squeeze-and-excitation block \cite{hu2018squeeze}.}



\begin{table}[t]
    \centering
    \caption{Specification for PSLT. ``conv2d'' and ``MBV2Block'' indicate the standard convolutional operation and the light-weight block in the MobileNetV2 \cite{sandler2018mobilenetv2} with the squeeze-and-excitation layer \cite{hu2018squeeze}. ``conv2d$\downarrow$'' and ``MBV2Block+SE$\downarrow$'' denote the operation or block for downsampling with stride 2. ``FC'' denotes the fully connected layer. \gjfour{PSLT has a total of 9.223M parameters according to the table.}}
    \setlength{\tabcolsep}{0.2mm}
    {\begin{tabular}{c|c|c|c|c|c}
        \hline
        Stage & Input & Block & \#Out & Stride & \#Params \\
        \hline
        Stem & \tabincell{c}{{$224^2 \times 3$} \\ {$112^2 \times 36$}} & \tabincell{c}{{$3\times 3$ conv2d$\downarrow$} \\ {$3\times 3$ conv2d $\times 2$}} & \tabincell{c}{{36}\\{36}} & \tabincell{c}{{2}\\{1}} & 0.018M\\
        \hline
        1 & \tabincell{c}{{$112^2 \times 36$} \\ {$56^2 \times 72$}} & \tabincell{c}{{MBV2Block+SE$\downarrow$} \\ {MBV2Block+SE $\times 2$}} & \tabincell{c}{{72}\\{72}} & \tabincell{c}{{2}\\{1}} & 0.125M \\
        \hline
        2 & \tabincell{c}{{$56^2 \times 72$} \\ {$28^2 \times 144$} \\ {$28^2 \times 144$}} & \tabincell{c}{{MBV2Block+SE$\downarrow$} \\ {MBV2Block+SE $\times 2$} \\{MBV2Block+SE$\downarrow$}} & \tabincell{c}{{144}\\{144}\\{288}} & \tabincell{c}{{2}\\{1}\\{2}} & 0.817M \\
        \hline
        3 & \tabincell{c}{{$14^2 \times 288$}} & \tabincell{c}{{Ladder SA block $\times 10$}} & \tabincell{c}{{288}} & \tabincell{c}{{1}} & 3.879M \\
        \hline
        4 & \tabincell{c}{{$14^2 \times 288$} \\ {$7^2 \times 576$}} & \tabincell{c}{{$2\times 2$ conv2d$\downarrow$} \\ {Ladder SA block $\times 3$}} & \tabincell{c}{{576}\\{576}} & \tabincell{c}{{2}\\{1}} & 3.808M \\
        \hline
        head & \tabincell{c}{{$7^2 \times 576$} \\ {$1^2 \times 576$}} & \tabincell{c}{{$7\times 7$, pool} \\ {FC}} & \tabincell{c}{{576}\\{1000}} & \tabincell{c}{{1}\\{1}} & 0.576M \\
        \hline
    \end{tabular}}
    \label{tab:architecture}
    \vspace{1.0cm}
\end{table}

\subsection{Network Specification}
Table \ref{tab:architecture} shows the architecture of PSLT, which contains 9.2M trainable parameters for image classification on the ImageNet dataset \cite{Russakovsky2015ImageNet}. PSLT stacks the light-weight convolutional blocks (MBV2Block+SE) in the first two stages and the proposed ladder self-attention blocks in the final two stages. PSLT starts with three $3\times 3$ convolution as stem to produce the input feature of the first stage. In both stage 1 and stage 2, the first convolutional block has a stride of 2 to downsample the input feature map. The final block in stage 2 is utilized to downsample the input feature map of stage 3. A $2\times 2$ convolution is applied to downsample the feature map of stage 4. Each time the size of the feature map is halved, the number of feature channels doubles.

Finally, the classifier head uses global average pooling on the output of the last block. The globally pooled feature passes through a fully connected layer to yield an image classification score. Without the classifier head, PSLT with multi-scale features can easily be applied as a backbone in existing methods for various visual tasks.

\vspace{0.2cm}

\noindent \textbf{- Detailed configuration.} \gjseven{Following  SwinTransformer \cite{liu2021swin}, the window size is set to $7\times 7$, and the window partition and position embedding is produced with the mechanism similar to \cite{liu2021swin}. The number of heads in PSW-MHSA is set to 4 and 8 in the last two stages respectively. The stride in the shift operation (shown in Figure \ref{fig:structure} (c)) in PSW-MHSA is set to 3. The shift operation is similar to that in \cite{liu2021swin}, but the shift direction is different from each other in the last two branches.}

\newpage

\section{Experiment}
In this section, we evaluate the effectiveness of the proposed PSLT by conducting experiments on several tasks, including ImageNet-1k image classification \cite{Russakovsky2015ImageNet}, COCO objection detection \cite{lin2014microsoft} and Market-1501 person re-identification \cite{zheng2015scalable}.

\subsection{Image Classification}

\subsubsection{ImageNet Results}
\label{sec:imagenet}

\noindent \textbf{- Experimental setting.} We evaluate our proposed PSLT on ImageNet \cite{Russakovsky2015ImageNet} classification. The ImageNet dataset has 1,000 classes, including 1.2 million images for training and 50 thousand images for validation. We train our PSLT on the training set, and the top-1 accuracy on the validation set \gjreone{and the ImageNet-V2 \cite{recht2019imagenet}} is reported. For a fair comparison, we follow Mobile-Former \cite{chen2021mobile} when performing the ImageNet classification experiments. The images were randomly cropped to $224 \times 224$. The data augmentation methods include horizontal flipping \cite{szegedy2015going}, mix-up \cite{zhang2017mixup}, auto-augmentation \cite{cubuk2019autoaugment} and random erasing \cite{zhong2020random}. Our PSLT was trained from scratch on 8 A6000 GPUs with label smoothing \cite{szegedy2016rethinking} using the AdamW \cite{loshchilov2017decoupled} optimizer for 450 epochs with a cosine learning rate decay. The distillation is not adopted for training. The initial learning rate was set to $6 \times 10^{-4}$, and the weight decay was set to 0.025 by default. For images in the validation set, we adopted the center crop, with images cropped to $224 \times 224$ for evaluation.

\begin{table*}[ht]
    \centering
    \caption{\gjreone{Image classification performance on the ImageNet without pretraining.} Models with similar number of parameters are presented for comparison. ``Input" indicates the scale of the input images. ``Top-1'' (``V2 Top-1'') denotes the top-1 accuracy on ImageNet (ImageNet-V2). ``\#Params'' refers to the number of trainable parameters. ``\#FLOPs'' is calculated according to the corresponding input size. ``+'' indicates that the performance is cited from Mobile-Former \cite{chen2021mobile}. ``*'' indicates that the model was trained with distilling from an external teacher. ``$\ddag$'' denotes extra techniques are applied, such as multi-scale sampler \cite{mehta2021mobilevit} and EMA \cite{Russakovsky2015ImageNet}. \gjreone{``PAcc'' (``FAcc'') is the ratio of Top-1 accuracy to the number of parameters (FLOPs).} }
    {\begin{tabular}{c|c|c|c|c|c|c|c|c}
        \hline
         Model & \gjretwo{NAS} & \tabincell{c}{Input} & \tabincell{c}{\#Params} & \#FLOPs &Top-1 & \tabincell{c}{PAcc (\%/M)}  & \tabincell{c}{FAcc (\%/G)} & V2 top-1 \\
        \hline
        \textbf{PSLT-Tiny(Ours)} & \xmark & $224^2$ & 4.3M & 876M & 74.9\% & 17.42 & 85.5 & 63.0\% \\
        \textbf{PSLT(Ours)} & \xmark & $224^2$ & 9.2M & 1.9G & 79.9\% & 8.68 & 42.05 & 68.6\% \\
        \gjreone{\textbf{PSLT-Large(Ours)}} & \xmark & $224^2$ & 16.0M & 3.4G & 81.5\% & 5.1 & 24.0 & 70.3\% \\
        \hline
        MobileNetV3 \cite{howard2019searching} & \cmark & $224^2$ & 4.0M & 155M & 73.3\% & 18.33 & 472.9 & -- \\
        EdgeViT-XXS \cite{pan2022edgevits} & \xmark & $224^2$ & 4.1M & 0.6G & 74.4\% & 18.14 & 124 & -- \\
        Mobile-Former \cite{chen2021mobile} & \xmark & $224^2$ & 4.6M & 96M & 72.8\% & 15.83 & 758.3 & -- \\
        $\textnormal{MobileViTv2-S \cite{mehta2021mobilevit}}^\ddag$ & \xmark & $256^2$ & 4.9M & 1.8G & 78.1\% & 15.94 & 43.39 & -- \\
        LVT \cite{yang2022lite} & \xmark & $224^2$ & 5.5M & 0.9G & 74.8\% & 13.6 & 83.11 & -- \\
        $\textnormal{MobileViT-S \cite{mehta2021mobilevit}}^\ddag$& \xmark & $256^2$ & 5.6M & 2.0G & 78.4\% & 14 & 39.2 & -- \\
        $\textnormal{EdgeNeXt-S \cite{maaz2022edgenext}}^\ddag$& \xmark & $224^2$ & 5.6M & 965M & 78.8\% & 14.07 & 81.66 & -- \\
        MPViT-T \cite{lee2021mpvit}& \xmark & $224^2$ & 5.8M & 1.6G & 78.2\% & 13.48 & 48.88 & -- \\
        HRFormer-T \cite{yuan2021hrformer}& \xmark & $224^2$ & 8.0M & 1.8G & 78.5\% & 9.81 & 43.61 & -- \\
        $\textnormal{LeViT \cite{graham2021levit}}^*$& \xmark & $224^2$ & 9.2M & 406M & 78.6\% & 8.54 & 193.6 & 66.6\% \\
        \gjreone{EfficientNet} \cite{tan2019efficientnet}& \cmark & $224^2$ & 9.2M & 1.0G & 80.1\% & 8.71 & 80.1 & 68.8\%\\
        RedNet-26 \cite{li2021involution}& \xmark & $224^2$ & 9.2M & 1.7G & 75.9\% & 8.25 & 44.65 & -- \\
        CMT-Ti \cite{guo2021cmt}& \xmark & $192^2$ & 9.5M & 0.6G & 79.2\% & 8.34 & 132 & -- \\
        ShuffleV2+Weight \cite{ma2020weightnet}& \xmark & $224^2$ & 9.6M & 307M & 75.0\% & 7.81 & 244.3 & -- \\
        \hline
        ConT-S \cite{yan2021contnet}& \xmark & $224^2$ & 10.1M & 1.5G & 76.5\% & 7.57 & 51 & -- \\
        Coat-Lite mini \cite{xu2021co}& \xmark & $224^2$ & 11.0M & 2.0G & 79.1\% & 7.19 & 39.55 & -- \\
        Shunted-T \cite{ren2021shunted}& \xmark & $224^2$ & 11.5M & 2.1G & 79.8\% & 6.94 & 38 & -- \\
        GC ViT-XXT \cite{hatamizadeh2022global}& \xmark & $224^2$ & 12.0M & 2.1G & 79.6\% & 6.63 & 37.9 & -- \\
        PoolFormer-S12 \cite{yu2021metaformer}& \xmark & $224^2$ & 12.0M & $\sim$ 4.2G & 77.2\% & 6.43 & 18.38 & -- \\
        EfficientFormer-L1 \cite{li2022efficientformer}& \cmark & $224^2$ & 12.2M & 2.4G & 79.2\% & 6.49 & 33 & -- \\
        $\textnormal{Swin-2G \cite{liu2021swin}}^+$& \xmark & $224^2$ & 12.8M & 2.0G & 79.2\% & 6.19 & 39.6 & -- \\
        PvT-Tiny \cite{wang2021pyramid}& \xmark & $224^2$ & 13.2M & 1.9G & 75.1\% & 5.69 & 39.53 & -- \\
        ResT-Small \cite{zhang2021rest}& \xmark & $224^2$ & 13.7M & 1.9G & 79.6\% & 5.81 & 41.89 & -- \\
        Mobile-Former \cite{chen2021mobile}& \xmark & $224^2$ & 14.0M & 508M & 79.3\% & 5.66 & 156.1 & -- \\
        \hline
        HRNet-W18 \cite{sun2019deep}& \xmark & $224^2$ & 21.3M & 4.0G & 76.8\% & 3.56 & 19.2 & -- \\
        PS-ViT-B/10 \cite{yue2021vision}& \xmark & $224^2$ & 21.3M & 3.1G & 80.6\% & 3.78 & 26 & -- \\
        A-ViT-S \cite{yin2022vit}& \xmark & $224^2$ & 22.0M & 3.6G & 78.6\% & 3.57 & 21.83 & -- \\
        $\textnormal{DeiT-S \cite{touvron2021training}}$& \xmark & $224^2$ & 22.0M & 4.6G & 79.8\% & 3.63 & 17.35 & 68.5\% \\
        BossNet-T0 \cite{li2021bossnas}& \cmark & $224^2$ & - - & 3.4G & 80.8\% & -- & 23.76 & -- \\
        PVT-S \cite{wang2021pyramid}& \xmark & $224^2$ & 24.5M & 3.8G & 79.8\% & 3.24 & 21 & -- \\
        Res2Net-50 \cite{gao2019res2net}& \xmark & $224^2$ & 25.0M & 4.2G & 78.0\% & 3.12 & 18.57 & -- \\
        ResNet-50 \cite{he2016deep}& \xmark & $224^2$ & 25.6M & 4.1G & 76.2\% & 2.98 & 18.59 & -- \\
        RedNet-101 \cite{li2021involution}& \xmark & $224^2$ & 25.6M & 4.7G & 79.1\% & 3.09 & 16.83 & -- \\
        VOLO \cite{yuan2021volo}& \xmark & $224^2$ & 27.0M & 6.8G & 84.2\% & 3.12 & 12.38 & -- \\
        CrossFormer-T \cite{wang2021crossformer}& \xmark & $224^2$ & 27.8M & 2.9G & 81.5\% & 2.93 & 28.1 & -- \\
        \gjreone{Swin} \cite{liu2021swin}& \xmark & $224^2$ & 29.0M & 4.5G & 81.3\% & 2.80 & 18.07 & --\\
        \gjretwo{WRN} \cite{zagoruyko2016wide}& \xmark & $224^2$ & 68.9M & -- & 78.1\% & 1.13 & -- & --\\
        $\textnormal{ViT-B/16 \cite{dosovitskiy2020image}} $& \xmark & $384^2$ & 86M & 55.5G & 77.9\% & 0.91 & 1.4 & 67.5\% \\
        \hline
    \end{tabular}}
    \label{tab:imagenet}
\end{table*}

\noindent \textbf{- Experimental results.} Table \ref{tab:imagenet} shows a comparison of the performance of our proposed PSLT and the performance of several convolution-based models and transformer-based models. The first block shows the performance of our proposed PSLT. The second block in Table \ref{tab:imagenet} presents the performance of models with less than 10M parameters, including light-weight convolutional neural networks and recently proposed transformer-based models. Compared to an efficient convolution-based model with a similar number of parameters (ShuffleNetV2 + WeightNet \cite{ma2020weightnet}), PSLT achieves a significantly higher top-1 accuracy (79.9\% vs. 75.0\%) with slightly fewer parameters (9.2M vs. 9.6M) but more FLOPs (1.9G vs. 307M). With a similar number of parameters (9.2M vs. 9.5M) and FLOPs (1.9G vs. 2.0G), PSLT achieves a top-1 accuracy that is 2\% higher than the pure transformer-based DeiT-2G (79.9\% vs. 77.6\%). Several recent works have focused on leveraging the advantages of CNNs and transformer-based architectures to improve model generalization ability and capacity, including LeViT \cite{graham2021levit} and CMT \cite{guo2021cmt}. It should be noted that our proposed PSLT (trained for 450 epochs without distillation) achieves a higher top-1 accuracy (79.9\% vs. 78.6\%) than LeViT (trained for 1,000 epochs with distillation) with a similar number of parameters. With a similar number of parameters (4.3M vs. 4.2M) and FLOPs (876M vs. 0.6G), PSLT-Tiny achieves a slightly higher top-1 accuracy (74.9\% vs. 74.4\%) than EdgeViT-XXS \cite{pan2022edgevits}. And PSLT achieves comparable performance to MPViT-T \cite{fan2021multiscale} with a similar number of FLOPs. 

\gjretwo{Although our PSLT shows slightly inferior performance compared to the EfficientNet (79.9\% vs. 80.1\%) with similar number of parameters, the EfficientNet is obtained by the carefully designed neural architecture search algorithm, which is time-consuming. And our PSLT achieves higher performance than the EfficientNet on the experiments of COCO for object detection in Table \ref{tab:COCO-detection} and instance segmentation in Table \ref{tab:COCO-instance}. Furthermore, our PSLT still achieves higher performance than EfficientFormer-L1 which automatically finds efficient transformers with similar neural architecture search algorithm to EfficientNet.}

\gjele{Compared to those recently developed lightweight networks \cite{pan2022edgevits, mehta2021mobilevit, maaz2022edgenext, lee2021mpvit, graham2021levit}, our PSLT provides a different perspective for acting as an effective backbone with a relatively small number of parameters and FLOPs. The above experimental results demonstrate that PSLT with diverse local self-attentions for long-range dependency is capable of achieving comparable performance, and modelling the local self-attention requires less computing resources than modelling the global self-attention.}

The third block in Table \ref{tab:imagenet} shows the performance of models with 10M $\sim$ 20M parameters. PSLT outperforms pure transformer-based architectures with a similar number of parameters and FLOPs (9.2M/1.9G), including PvT-Tiny (13.2M/1.9G) and DeiT-2G (6.5M/2.0G), and PSLT achieves a slightly higher top-1 accuracy than Swin-2G (79.9\% vs. 79.2\%) with slightly fewer parameters. PSLT slightly outperforms CNN-based models and transformer-based models, where PSLT achieves a higher top-1 accuracy than Mobile-Former (79.9\% vs. 79.3\%) with significantly fewer parameters (9.2M vs. 14.0M) and more FLOPs (1.9G vs. 508M), and PSLT achieves a top-1 accuracy that is 3\% higher than ConT-S (79.9\% vs. 76.5\%) with a similar number of parameters and FLOPs. \gjtwleve{PSLT also achieves comparable performance to the searched light-weight model EfficientFormer-L1 \cite{li2022efficientformer} (79.9\% vs. 79.2\%), where PSLT requires a relatively smaller number of parameters (9.2M vs. 12.2M) and FLOPs (1.9G vs. 2.4G).}

The last block in Table \ref{tab:imagenet} shows the performance of models with more than 20M parameters for reference, including the commonly used ResNet and its evolved architectures. 
Compared to the pure transformer models, PSLT achieves a comparable top-1 accuracy to PVT-S (79.9\% vs. 79.8\%) with nearly 50\% fewer parameters (9.2M vs. 24.5M) and FLOPs (1.9G vs. 3.8G). \gjreone{PSLT-Large with smaller amount of parameters and FLOPs (16.0M/3.4G) also outperforms recent transformer structures, including DeiT-S (22.0M/4.6G), PS-ViT-B/10 (21.3M/3.1G) and BossNet-T0 (3.4G). And the PSLT-Large also achieves a slightly higher top-1 accuracy than Swin \cite{liu2021swin} (81.5\% vs. 81.3\%) with fewer parameters (16.0M vs. 29.0M) and FLOPs (3.4G vs. 4.5G).}



The ImageNet classification experiments show that our proposed PSLT outperforms both convolution-based and transformer-based models with similar number of parameters. These results validate the effectiveness of the proposed ladder self-attention block.

\subsubsection{CIFAR Results}

\vspace{0.1cm}

\noindent \textbf{- Experimental setting.} The CIFAR-10/100 \cite{krizhevsky2009learning} datasets contain 10/100 classes, including 50,000 images for training and 10,000 images for testing, with an image resolution of $32\times32$. We trained PSLT from scratch on the training set, and the top-1 accuracy on the test set is reported. PSLT was trained for 300 epochs using the AdamW \cite{loshchilov2017decoupled} optimizer with cosine learning rate decay using an initial learning rate of $5 \times 10^{-4}$ and a weight decay of 0.025.

\begin{table}[t]
    \centering
    \caption{Image classification performance of the models on the CIFAR-10/100 without pretraining. The top-1 accuracy rate on CIFAR-10/100 is reported. ``+'' denotes that results were taken from \cite{patel2022aggregating}. \gjthirteen{``FLOPs'' is not reported for the compared methods in literatures.}}
    \setlength{\tabcolsep}{1.0mm}
    {\begin{tabular}{c|c|c|c|c}
        \hline
        Model& \tabincell{c}{Input} & \tabincell{c}{\#Params}  & \tabincell{c}{CIFAR-10} & CIFAR-100 \\
        \hline
        \gjreone{ResNet} \cite{he2016deep} & $32^2$ & 19.4M & 92.07\% & -- \\
        \gjreone{WRN-40-4} \cite{zagoruyko2016wide} & $32^2$ & 8.9M & 95.47\% & 78.82\% \\
        \gjreone{ResNeXt-29} \cite{xie2017aggregated} & $32^2$ & 68.1M & 96.43\% & 82.69\% \\
        \gjreone{DenseNet} \cite{huang2017densely} & $32^2$ & 27.2M & 94.17\% & 76.58\% \\
        \hline
        $\textnormal{PVT-T \cite{wang2021pyramid}}^+$ & $32^2$ & 13M & 91.0\% & 72.80\% \\
        $\textnormal{Swin \cite{liu2021swin}}^+$ & $32^2$ & 27.5M & 94.41\% & 78.07\% \\
        MOA-T \cite{patel2022aggregating} & $32^2$ & 30M & 95.0\% & 78.63\% \\
        \hline
        \textbf{PSLT (Ours)} & $32^2$  & 8.7M & 95.43\% & 79.1\% \\
        \hline
    \end{tabular}}
    \label{tab:CIFAR}
\end{table}

\vspace{0.1cm}

\begin{table*}[ht]
    \centering
    \caption{Object detection performance on COCO. The $AP$ value on the validation set is reported. ``MS'' denotes that multi-scale training \cite{wang2021pyramid} was used. ``+'' indicates that the result is taken from \cite{wang2021pyramid}. FLOPs is calculated at input resolution $1280\times 800$. \gjreone{``*'' denotes that the model is trained based on Sparse R-CNN \cite{sun2021sparse} following \cite{liu2021swin}.}}
    \setlength{\tabcolsep}{1mm}
    {\begin{tabular}{c|c|c|ccc|ccc|ccc|ccc}
        \hline
        \multirow{2}{*}{Backbone} & \multirow{2}{*}{\#Params} & \multirow{2}{*}{\#FLOPs}  & \multicolumn{6}{c|}{RetinaNet 1x} & \multicolumn{6}{c}{RetinaNet 3x + MS} \\
        \cline{4-15} & & & $AP$ & $AP_{50}$ & $AP_{75}$ & $AP_S$ & $AP_M$ & $AP_L$ & $AP$ & $AP_{50}$ & $AP_{75}$ & $AP_S$ & $AP_M$ & $AP_L$ \\
        \hline 
        \textbf{PSLT(Ours)} & 19.1M & 192G & 41.2 & 61.3 & 44.0 & 25.5 & 45.0 & 54.8 & 43.9 & 64.3 & 46.9 & 27.6 & 47.6 & 56.9  \\
        \hline
        $\textnormal{Mobile-Former \cite{chen2021mobile}}$ & 17.9M & 181G & 38.0 & 58.3 & 40.3 & 22.9 & 41.2 & 49.7 & -- & -- & -- & -- & -- & -- \\
        \gjreone{EfficientNet \cite{tan2019efficientnet}} & 20.0M & 151.2G & 40.5 & 60.5 & 43.1 & 22.2 & 44.8 & 56.6 & 42.4 & 62.0 & 45.6 & 24.1 & 46.1 & 58.3 \\ 
        $\textnormal{ResNet-18 \cite{he2016deep}}^+$ & 21.3M & 168G & 31.8 & 49.6 & 33.6 & 16.3 & 34.2 & 43.2 & 35.4 & 53.9 & 37.6 & 19.5 & 38.2 & 46.8 \\
        $\textnormal{PVT-T \cite{wang2021pyramid}}$ & 23.0M & 221G& 36.7 & 56.9 & 38.9 & 22.6 & 38.8 & 50.0 & 39.4 & 59.8 & 42.0 & 25.5 & 42.0 & 52.1 \\
        RedNet-50 \cite{li2021involution} & 27.8M & 210G & 38.3 & 58.2 & 40.5 & 21.1 & 41.8 & 50.9 & -- & -- & -- & -- & -- & -- \\
        $\textnormal{ConT-M \cite{yan2021contnet}}$ & 27.0M & 217G & 37.9 & 58.1 & 40.2 & 23.0 & 40.6 & 50.4 & -- & -- & -- & -- & -- & -- \\
        \hline
        $\textnormal{PVT-S \cite{wang2021pyramid}}$ & 34.2M & 226G & 40.4 & 61.3 & 43.0 & 25.0 & 42.9 & 55.7 & 42.2 & 62.7 & 45.0 & 26.2 & 45.2 & 57.2 \\
        $\textnormal{ResNet-50 \cite{he2016deep}}^+$ & 37.7M & 239G & 36.3 & 55.3 & 38.6 & 19.3 & 40.0 & 48.8 & 39.0 & 58.4 & 41.8 & 22.4 & 42.8 & 51.6 \\
        \hline
        \gjreone{\textbf{PSLT-Large (Ours)$^*$}} & 97.6M & 147.7G & -- & -- & -- & -- & -- & -- & 48.2 & 67.2 & 52.8 & 32.3 & 51.1 & 62.4  \\
        \gjreone{Swin \cite{liu2021swin}$^*$} & 110.0M & 172.0G & -- & -- & -- & -- & -- & -- & 47.9 & 67.3 & 52.3 & -- & -- & --  \\
        \hline
    \end{tabular}
    }
    \label{tab:COCO-detection}
\end{table*}

\begin{table*}[ht]
    \centering
    \caption{Object detection and instance segmentation performance on COCO. $AP^b$ and $AP^m$ denote the bounding box AP and mask AP on the validation set, respectively. ``MS'' denotes that multi-scale training \cite{wang2021pyramid} was used. ``+'' indicates that the result was taken from \cite{wang2021pyramid}. FLOPs is calculated at input resolution $1280\times 800$. }
    \setlength{\tabcolsep}{1mm}
    {\begin{tabular}{c|c|c|ccc|ccc|ccc|ccc}
        \hline
        \multirow{2}{*}{Backbone} & \multirow{2}{*}{\#Params} & \multirow{2}{*}{\#FLOPs}  & \multicolumn{6}{c|}{Mask R-CNN 1x} & \multicolumn{6}{c}{Mask R-CNN 3x + MS} \\
        \cline{4-15} & & & $AP^b$ & $AP^b_{50}$ & $AP^b_{75}$ & $AP^m$ & $AP^m_{50}$ & $AP^m_{75}$ & $AP^b$ & $AP^b_{50}$ & $AP^b_{75}$ & $AP^m$ & $AP^m_{50}$ & $AP^m_{75}$ \\
        \hline 
        \textbf{PSLT (Ours)} & 28.9M & 211G & 40.8 & 61.8 & 44.9 & 37.3 & 59.0 & 39.9 & 42.7 & 63.1 & 47.1 & 38.9 & 60.5 & 41.9 \\
        \hline
        \gjreone{EfficientNet} \cite{tan2019efficientnet} & 30.2M & 169.3G & 40.1 & 61.2 & 43.5 & 36.4 & 58.0 & 38.9 & 42.1 & 63.3 & 45.5 & 38.0 & 60.2 & 40.6 \\
        RedNet-50 \cite{li2021involution} & 34.2M & 224G & 40.2 & 61.4 & 43.7 & 36.1 & 58.1 & 38.2 & -- & -- & -- & -- & -- & -- \\
        $\textnormal{ResNet-18 \cite{he2016deep}}^+$ & 31.2M & --  --& 34.0 & 54.0 & 36.7 & 31.2 & 51.0 & 32.7 & 36.9 & 57.1 & 40.0 & 33.6 & 53.9 & 35.7 \\
        $\textnormal{PVT-T \cite{wang2021pyramid}}$ & 32.9M & 240G & 36.7 & 59.2 & 39.3 & 35.1 & 56.7 & 37.3 & 39.8 & 62.2 & 43.0 & 37.4 & 59.8 & 39.9 \\
        \hline
        $\textnormal{PVT-S \cite{wang2021pyramid}}$ & 44.1M & 305G & 40.4 & 62.9 & 43.8 & 37.8 & 60.1 & 40.3 & 43.0 & 65.3 & 46.9 & 39.9 & 62.5 & 42.8 \\
        $\textnormal{ResNet-50 \cite{he2016deep}}^+$ & 44.2M & 253G & 38.0 & 58.6 & 41.4 & 34.4 & 55.1 & 36.7 & 41.0 & 61.7 & 44.9 & 37.1 & 58.4 & 40.1 \\
        \hline
        \gjreone{\textbf{PSLT-Large (Ours)}} & 35.6M & 242.3G & 44.1 & 65.9 & 48.3 & 39.6 & 62.6 & 42.4 & 46.7 & 68.0 & 51.5 & 41.8 & 65.0 & 45.2 \\
        \gjreone{Swin \cite{liu2021swin}} & 48.0M & 267.0G & 43.7 & 66.6 & 47.7 & 39.8 & 63.3 & 42.7 & 46.0 & 68.1 & 50.3 & 41.6 & 65.1 & 44.9 \\
        \hline
    \end{tabular}
    }
    \label{tab:COCO-instance}
\end{table*}

\noindent\textbf{- Experimental results.} Table \ref{tab:CIFAR} shows the image classification performance of the vision transformer models on the CIFAR-10 and CIFAR-100 datasets. \gjreone{As shown in the table, our PSLT achieves comparable performance compared to the commonly-used convolutional neural networks with a small amount of parameters. \gjretwo{Compared with WRN \cite{zagoruyko2016wide}, PSLT achieves comparable performance on the CIFAR10/100 with similar amount of parameters but higher top-1 accuracy on ImageNet with much less parameters in Table \ref{tab:imagenet}.} Compared to the vision transformers, our PSLT shows higher model generalization ability with fewer training images, where PSLT achieves relatively higher top-1 accuracy on both CIFAR-10 (95.43\% vs. 95.0\%) and CIFAR-100 (79.1\% vs. 78.63\%) than MOA-T with less trainable parameters (8.7M vs. 30M).}

\subsection{Objection Detection}

\noindent \textbf{- Experimental setting.} We conduct objection detection experiments on COCO 2017 \cite{lin2014microsoft}, which contains 118k images in the training set and 5k images in the validation set of 80 classes. Following PVT \cite{wang2021pyramid}, which uses standard settings to generate multi-scale feature maps, we evaluate our proposed PSLT as a backbone with two standard detectors: RetinaNet \cite{lin2017focal} (one-stage) and Mask R-CNN \cite{he2017mask} (two-stage). During training, our backbone (PSLT) was first initialized with the pretrained weights on ImageNet \cite{Russakovsky2015ImageNet}, and the newly added layers were initialized with Xavier \cite{glorot2010understanding}. Our PSLT was trained with a batch size of 16 on 8 GeForce 1080Ti GPUs. We adopt the AdamW \cite{loshchilov2017decoupled} optimizer with an initial learning rate of $1 \times 10^{-4}$. Following the common PVT settings, we adopted a $1 \times$ or $3 \times$ training schedule (12 or 36 epochs) to train the detection models. The shorter side of the training images was resized to 800 pixels, while the longer side of the images is set to a maximum of 1,333 pixels. The shorter side of the validation images was fixed to 800 pixels. With the $3 \times$ training schedule, the shorter side of the input image was randomly resized in the range of [640, 800].

\vspace{0.1cm}

\noindent\textbf{- Experimental results.} Table \ref{tab:COCO-detection} shows the performance of the backbones on COCO val2017; the backbones were initialized with pretrained weights on ImageNet using RetinaNet for object detection. As shown in the table, with a similar number of trainable parameters, PSLT achieves comparable performance. With the $1 \times$ training scheduler, PSLT outperforms Mobile-Former, where PSLT achieving a higher AP (41.2 vs. 38.0) with slightly more parameters (19.1 M vs. 17.9 M) and FLOPs (192 G vs. 181G). The last block of Table \ref{tab:COCO-detection} presents models with more than 30M parameters; PSLT achieves higher AP (41.2 vs. 40.4) than PVT-S with considerably fewer parameters (19.1 M vs. 34.2M) and FLOPs (192 G vs. 226 G). Compared with the commonly used backbone ResNet-50, PSLT achieves a considerably higher AP (41.2 vs. 35.3) with nearly 50\% less trainable parameters (19.1 M vs. 37.7 M) and fewer FLOPs (192 G vs. 239 G). With the $3 \times$ training scheduler, PSLT shows higher performance with a similar number of parameters, where PSLT achieving a 1.7-point higher AP than PVT-S (43.9 vs. 42.2) with considerably fewer parameters.

Similar results are shown in Table \ref{tab:COCO-instance}, including segmentation experiments on Mask R-CNN. With the $1 \times$ training scheduler, PSLT outperforms ResNet-50, where PSLT achieving 2.8 points higher box AP (40.8 vs. 38.0) and 2.9 points higher mask AP (37.3 vs. 34.4) with fewer parameters (28.9M vs. 44.2M) and FLOPs (211 G vs. 253 G). PSLT shows comparable performance to PVT-S (40.8 vs. 40.4 box AP and 37.3 vs. 37.8 mask AP) with considerably fewer parameters (28.9M vs. 44.1M) and FLOPs (211G vs. 305G). Similar results are observed with the $3 \times$ training scheduler; PSLT outperforms PVT-T, where PSLT achieving 2.9 points higher $AP^b$ and 1.5 points higher $AP^m$ with a smaller number of parameters (28.9M vs. 32.9M) and FLOPs (211 G vs. 240 G). Moreover, PSLT achieves comparable performance to ResNet-50, which has more than 40M parameters.

\gjreone{For a fair comparison with Swin \cite{liu2021swin}, we implement our PSLT-Large with the same experimental setting as Swin \cite{liu2021swin} with Sparse R-CNN \cite{sun2021sparse} framework (in Table \ref{tab:COCO-detection}) and with Mask R-CNN \cite{he2017mask} framework (in Table \ref{tab:COCO-instance}). Our PSLT-Large achieves comparable performance with less parameters and FLOPs.}


\subsection{Semantic Segmentation}

\noindent \textbf{- Experimental setting.} Semantic segmentation experiments are conducted with a challenging scene parsing dataset, ADE20K \cite{zhou2017scene}. ADE20K contains 150 fine-grained semantic categories with 20210, 2000 and 3352 images in the training, validation and test sets, respectively. Following PVT \cite{wang2021pyramid}, we evaluate our proposed PSLT as a backbone on the basis of the Semantic FPN \cite{kirillov2019panoptic}, a simple segmentation method without dilated convolution \cite{yu2015multi}. In the training phase, the backbone was first initialized with the pretrained weights on ImageNet \cite{Russakovsky2015ImageNet}, and the newly added layers were initialized with Xavier \cite{glorot2010understanding}. Our PSLT was trained for 80k iterations with a batch size of 16 on 4 GeForce 1080Ti GPUs. We adopted the AdamW \cite{loshchilov2017decoupled} optimizer with an initial learning rate of $2 \times 10^{-4}$. The learning rate follows the polynomial decay rate with a power of 0.9. The training images were randomly resized and cropped to 512 $\times$ 512, and the images in the validation set were rescaled, with the shorter side set to 512 pixels during testing.

\noindent \textbf{- Experimental results.} Table \ref{tab:segmentation} shows the semantic segmentation performance on the ADE20K validation set for backbones initialized with pretrained weights on ImageNet with Sematic FPN \cite{kirillov2019panoptic}. As shown in the table, PSLT outperforms the ResNet-based models \cite{he2016deep}, where PSLT achieving a 
2.3\% points higher than ResNet-50 (39.0\% vs. 36.7\%) with nearly half the number of parameters (13.0M vs. 28.5M) and 20\% less FLOPs than ResNet-50 (32.0G vs. 45.6G). With almost the same number parameters and FLOPs, PSLT achieves a 
2.3\% points higher than PVT-T (39.0\% vs 36.7\%). 

\gjreone{For a fair comparison with Swin \cite{liu2021swin}, we implement our PSLT-Large with the same experimental setting as Swin \cite{liu2021swin} with UperNet \cite{xiao2018unified} framework. Our PSLT-Large achieves comparable performance with less parameters and FLOPs.}

\begin{table}[t]
    \centering
    \caption{Semantic segmentation performance of different backbones on the ADE20K. The mean intersection over union (MIoU) is reported. ``+'' indicates that the result is taken from \cite{wang2021pyramid}. \gjreone{``*'' denotes that the model is trained based on UperNet \cite{xiao2018unified} following \cite{liu2021swin}.}}
    \setlength{\tabcolsep}{1.7mm}
    {\begin{tabular}{c|cccc}
        \hline
        \multirow{2}{*}{Backbone}  & \multicolumn{4}{c}{Semantic FPN} \\
        \cline{2-5} & \#Params & Input & FLOPs & mIoU \\
        \hline 
        \textbf{PSLT (Ours)} & 13.0M & $512^2$ & 32.0G & 39.0\% \\
        \hline
        $\textnormal{ResNet-18 \cite{he2016deep}}^+$ & 15.5M & $512^2$ & 32.2G & 32.9\% \\
        $\textnormal{PVT-T \cite{wang2021pyramid}}$ & 17.0M & $512^2$ & 33.2G & 35.7\% \\
        \hline
        $\textnormal{ResNet-50 \cite{he2016deep}}^+$ & 28.5M & $512^2$ & 45.6G & 36.7\% \\
        $\textnormal{PVT-S \cite{wang2021pyramid}}$ & 28.2M & $512^2$ & 44.5G & 39.8\%  \\
        \hline
        \gjreone{\textbf{PSLT-Large(Ours)$^*$}} & 47.8M & $512^2$ & 229.9G & 46.1\%\\
        \gjreone{Swin\cite{liu2021swin}$^*$} & 60M & $512^2$ & 945G & 46.1\% \\
        
        \hline
    \end{tabular}
    }
    \label{tab:segmentation}
\end{table}

\subsection{Person Re-identification}

\vspace{0.1cm}

\noindent \textbf{- Experimental setting.} Person re-identification (ReID) experiments are conducted with Market-1501 \cite{zheng2015scalable}, a dataset that contains 12,936 images of 751 people in the training set and 19,732 images of 750 different people in the validation set. All the images in the Market-1501 dataset were captured with six cameras. We trained our PSLT on the training set to classify each person, similar to an image classifier with 751 training classes. The output feature of the final stage in PSLT was utilized to compute the similarity between the query image and the gallery images for validation, and the rank-1 accuracy and mean average precision (mAP) are reported. For a fair comparison, we follow Bag of Tricks \cite{luo2019bag} to perform the ReID experiments. The data augmentation techniques that we apply to the training images include horizontal flipping, central padding, randomly cropping to $256 \times 128$, normalizing by subtracting the channel mean and dividing by the channel standard deviation, and random erasing \cite{zhong2020random}. The data augmentation techniques applied to the test images include resizing to $256 \times 128$ and normalizing by subtracting the channel mean and dividing by the channel standard deviation.

\begin{table}[t]
    \centering
    \caption{Performance of the models on the Market-1501 when the models are first pretrained on ImageNet. ``Rank-1'' indicates the rank-1 accuracy rate. ``mAP'' denotes the mean Average Precision. \gjthirteen{``FLOPs'' is not reported for the compared methods in literatures.}}
    \setlength{\tabcolsep}{1.0mm}
    {\begin{tabular}{c|c|c|c|c}
        \hline
        Model& \tabincell{c}{Input} & \tabincell{c}{\#Params}  & \tabincell{c}{Rank-1} & mAP \\
        \hline 
        OSNet \cite{zhou2019omni} & $256 \times 128$ & 2.2M & 94.8\% & 84.9 \\
        CDNet \cite{li2021combined} & $256 \times 128$ & 1.8M & 95.1\% & 86.0 \\
        Auto-ReID \cite{quan2019auto} & $256 \times 128$ & 11.4M & 94.5\% & 85.1 \\
        TransReID \cite{he2021transreid}  & $256 \times 128$ & $\sim$50M & 94.7\% & 86.8 \\
        \hline
        PCB (+RPP) \cite{sun2018beyond} & $256 \times 128$ & $\sim$26M & 93.8\% & 81.6 \\
        VPM \cite{sun2019perceive} & $256 \times 128$ & $\sim$26M & 93.0\% & 80.8 \\
        $\textnormal{BagofTrick \cite{luo2019bag}}$ & $256 \times 128$ & $\sim$26M & 94.5\% & 85.9 \\
        \textbf{PSLT (Ours)} & $256 \times 128$  & 9.0M & 94.3\% & 86.2 \\
        \hline
    \end{tabular}}
    \label{tab:market-pretrain}
\end{table}

The proposed PSLT was trained for 360 epochs under the guidance of the cross-entropy loss and triplet loss \cite{hermans2017in} using the Adam optimizer. Similarly, PSLT adopts the BNNeck \cite{luo2019bag} structure to compute the triplet loss. The output feature after the global average pooling layer first passed through a BatchNorm layer before being sent to the classifier. The feature before the BatchNorm layer was used to compute the triplet loss, and the output score of the classifier was used to compute the cross-entropy loss for training. For validation, the feature after the BatchNorm layer was adopted to compute the similarity. A batch size of 64 was applied during training. The initial learning rate is set to $1.5 \times 10^{-4}$, and the learning rate was decayed by a factor of 10 at epochs 150, 225 and 300. The weight decay was set to $5 \times 10^{-4}$.

\vspace{0.1cm}

\noindent \textbf{- Experimental results.} Table \ref{tab:market-pretrain} shows the performance on Market-1501 of models that were pretrained on ImageNet. The models in the first block are designed specifically for person ReID. The second block presents the performance of general-purpose backbones with improvement methodologies on Market-1501. PSLT outperforms some methods with general-purpose backbones. PSLT achieves a higher rank-1 accuracy (94.3\% vs. 93.8\%) and mAP (86.2 vs. 81.6) than PCB (+RPP) with fewer parameters (9.0M vs. 26M). In addition, PSLT achieves comparable performance to Auto-ReID, which is automatically designed for person ReID. Compared to Auto-ReID, PSLT achieves a higher mAP (86.2 vs. 85.1) and the same rank-1 accuracy with fewer parameters (9.0M vs. 11.4M). PSLT also shows comparable performance to OSNet, with a slightly higher mAP (86.2 vs. 84.9) and slightly lower rank-1 accuracy (94.3\% vs. 94.8\%). However, PSLT performs worse than specifically designed state-of-the-art models (CDNet and TransReID).

As can be seen, our PSLT achieves comparable performance to general-purpose models in the second block and slightly inferior performance to the models specifically designed for ReID in the first block. The experimental results demonstrate that our PSLT is generalizable to the person re-identification tasks.

        

\begin{table}[t]
    \centering
    \caption{\gjreone{Comparison of model inference. ``Mem'' denotes the peak memory for evaluation. ``FPS'' is the number of images processed for one second.}}
    \setlength{\tabcolsep}{1mm}
    {\begin{tabular}{c|c|c|c|c|c}
        \hline
         Model & \#Params & \#FLOPs & Mem & FPS &Top-1 \\
        \hline
        \textbf{PSLT-Tiny}  & 4.3M & 876M & 1.9G & 669.2  & 74.9\% \\
        \textbf{PSLT} & 9.2M & 1.9G & 2.1G & 638.2 & 79.9\% \\
        \gjreone{\textbf{PSLT-Large}} & 16.0M & 3.4G & 2.3G & 557.7 & 81.5\% \\
        \hline
        $\textnormal{DeiT-S \cite{touvron2021training}}$ & 22.0M  & 4.6G & 1.8G  & 1456.9 & 79.8\% \\
        \gjreone{Swin} \cite{liu2021swin} & 29.0M & 4.5G & 2.5G & 755.2 & 81.3\% \\
        \gjreone{Swin$^{\times3}$} \cite{liu2021swin} & 247.2M & 38.1G & 4.8G & 272.4 & -- \\
        
        \hline
    \end{tabular}}
    \label{tab:model-inference}
\end{table}

\subsection{Discussion}
\gjreone{We have shown our model performance in Table \ref{tab:imagenet} and our PSLT achieves comparable top-1 accuracy with relative less parameters and FLOPs. Usually, the FLOPs is adopted to measure the total float operations for processing one image. Intuitively, the model with lower FLOPs can achieve higher FPS which measures the amount of processed images in one second. However, according to Table \ref{tab:model-inference}, the FPS for the listed three kinds of models seems counter-intuitive. For example, the Swin contains relatively less FLOPs than DeiT-S but processes nearly a half amount of images as the DeiT-S does in one second. The same phenomenon occurs when comparing our PSLT-Large and Swin. The main reason is that the computing optimization is not implemented for Swin and our PSLT-Large directly on the accelerators such as GPUs, including the window partition in Swin and PSLT-Large and the computation of multiple branches in our PSLT-Large. Furthermore, we examine the FPS of Swin$^{\times3}$ with 3 times of the original channel dimension, and find that the FLOPs is nearly 9 times of the Swin but the FPS is only 1/3 of the Swin. Thus, more direct optimization on the computing hardware (GPU) for each transformer model could essential for the FPS. However, the optimization of the computing hardware for fast inference can be investigated in the future but is not the main scope of our work.
}

\section{Ablation Study}
In this section, we conduct ImageNet classification experiments to demonstrate that the ladder self-attention block and progressive shift mechanism in the proposed PSLT are effective. Here, all models were trained with an input image resolution of $224^2$ following the experimental settings described in Section \ref{sec:imagenet}.

\begin{table}[t]
    \centering
    \caption{Ablation study on the ImageNet without pretraining. ``\#Branches'' indicates the number of branches in the ladder self-attention block.}
    {\begin{tabular}{c|c|c|c|c}
        \hline
         Model& \#Branches & \tabincell{c}{\#Params} & \#FLOPs &Top-1 \\
        \hline
        PSLT & 2 & 11.0M & 2.1G & 80.2\% \\
        PSLT & 3 & 9.2M & 1.9G & 79.9\% \\
        PSLT & 4 & 8.4M & 1.8G & 79.0\% \\
        PSLT & 6 & 7.5M & 1.7G & 77.9\% \\
        
        \hline
    \end{tabular}}
    \label{tab:ablation-branch}
\end{table}

\vspace{0.1cm}

\noindent \textbf{- Number of branches in the ladder self-attention block.} \gjsix{As shown in Section \ref{sec:multi-branch}, the ladder self-attention block contains multiple branches, and more branches denotes that the developed backbone has less trainable parameters. In the ladder self-attention block with 6 branches, the shift stride in each branch is set to 3 (with two directions), 1 (with two directions), 5. The blocks with less branches are constructed similarly to the above. Table \ref{tab:ablation-branch} shows the performance of PSLT with various number of branches in the ladder self-attention blocks. Although the model with more branches requires less computing resources, the performance of the model gets worse. For balance, we set the number of branches as 3 by default.} 

\begin{table}[t]
    \centering
    \caption{Ablation study on the ImageNet without pretraining. ``Shift'' denotes that the shift operation was used in the ladder self-attention block to model diverse local self-attentions. ``FD'' indicates whether progressively transmitting features is adopted. ``PAFM'' indicates the pixel-adaptive fusion module.}
    \setlength{\tabcolsep}{1.5mm}{\begin{tabular}{c|c|c|c|c|c|c}
        \hline
         Model& Shift & FD & PAFM & \tabincell{c}{\#Params} & \#FLOPs &Top-1 \\
        \hline
        PSLT & \cmark & \cmark & \cmark & 9.2M & 1.9G & 79.9\% \\
        \hline
        -- -- & \xmark & \cmark & \cmark & 9.2M & 1.9G & 79.3\% \\
        -- -- & \cmark & \xmark & \cmark & 9.7M & 2.0G & 79.2\% \\
        \hline
    \end{tabular}}
    \label{tab:ablation}
\end{table}

\vspace{0.1cm}

\noindent \textbf{- Shift operation in the ladder self-attention block.} As described in Section \ref{sec:progressive}, PSLT adopts the shift operation in each branch of the ladder self-attention block to model long-range interactions. The third line in Table \ref{tab:ablation} shows the performance of PSLT without the shift operation, \textit{i.e.}, all shift operations in Figures \ref{fig:structure} and \ref{fig:multi-branch-SA} are removed. The experimental results show that with the shift operation, PSLT achieves a higher top-1 accuracy (79.9\% vs. 79.3\%) without increasing the number of training parameters or FLOPs. 

\vspace{0.1cm}

\noindent \textbf{- Feature delivery in the ladder self-attention block.} For long-range interactions, PSLT progressively delivers features of previous branch, where the latter branch takes the output features of the previous branch as its value when computing self-attention. Here, we investigate whether delivering the output features is effective. A multi-branch block is adopted, and the shift operation is still used for each branch, \ie only the horizontal connections in Figure \ref{fig:structure} are removed, and the PSW-MHSA in each branch is replaced with the W-MHSA in Figure \ref{fig:multi-branch-SA}. The last line in Table \ref{tab:ablation} presents the performance of the model without horizontal delivery. Since the value of self-attention in each branch needs to be computed from the input feature, the number parameters in the model without horizontal delivery is larger than the PSLT. With the horizontal connections in the ladder self-attention block, PSLT achieves a higher top-1 accuracy (79.9\% vs. 79.2\%) with slightly fewer parameters (9.2M vs. 9.7M).

\gjeight{By modelling diverse local interactions and interacting among the branches, the progressive shift mechanism is capable of enlarging the receptive field of the ladder self-attention block. The last two lines in Table \ref{tab:ablation} show performance of models without the two strategies respectively. Obviously, PSLT with the progressive shift mechanism achieves higher performance. Besides, with similar amount of computing resources, PSLT also outperforms the Swin-2G as shown in Table \ref{tab:imagenet}, where Swin-2G is implemented in the same way as SwinTransformer \cite{liu2021swin}.} \gjseven{The above experimental results show that the progressive shift mechanism improves model capacity with large receptive field.}

\begin{table}[t]
    \centering
    \caption{Ablation study of FFN layer on ImageNet without pre-training. ``LFFN'' denotes our Light FNN layer.}
    {\begin{tabular}{c|c|c|c|c}
        \hline
         Model& FFN & \tabincell{c}{\#Params} & \#FLOPs &Top-1 \\
        \hline
        PSLT & LFFN & 9.2M & 1.9G & 79.9\% \\
        -- -- & FFN & 12.8M & 2.3G & 79.7\% \\
        \hline
    \end{tabular}}
    \label{tab:ablation-FFN}
\end{table}

\begin{table}[t]
    \centering
    \caption{Ablation study of the fusion module for the multiple branches on the ImageNet without pretraining. \gjten{``Module'' indicates the method for feature fusion. ``Weight'' denotes the existence of the adaptive weights for feature fusion. ``Pointwise'' indicates the existence of a fully connected layer for fusing the features along the channel dimension. ``AW'' denotes that only the first two fully connected layers are preserved in the pixel-adaptive fusion module. ``Concat'' indicates that the output features of the multiple branches are concatenated}. ``SE'' denotes that the SE \cite{hu2018squeeze} block is adopted.}
    \setlength{\tabcolsep}{1mm}{\begin{tabular}{c|c|c|c|c|c}
        \hline
         Module & Weight & Pointwise & \tabincell{c}{\#Params} & \#FLOPs &Top-1 \\
        \hline
        PAFM & \cmark & \cmark & 9.2M & 1.9G & 79.9\% \\
        \hline
        FC & \xmark & \cmark & 7.4M & 1.7G & 78.8\% \\
        AW & \cmark & \xmark & 7.4M & 1.7G & 78.6\% \\
        Concat & \xmark & \xmark & 5.6M & 1.5G & 77.3\%\\
        \hline
        SE \cite{hu2018squeeze} & \cmark & \cmark & 9.2M & 1.7G & 79.0\% \\
        \hline
    \end{tabular}}
    \label{tab:ablation-PAFM}
\end{table}

\vspace{0.1cm}

\noindent \textbf{- Light FFN layer.} \gjeight{For further cutting down the computing budgets, PSLT proposes the Light FNN (LFFN) layer to replace the original FFN layer in self-attention block. Table \ref{tab:ablation-FFN} shows the comparison between PSLT and the model with FFN (only Light FNN layer in PSLT is replaced with the original FFN layer), and PSLT achieves comparable performance, where PSLT with LFFN achieves comparable top-1 accuracay (79.9\% vs. 79.7\%) with relatively smaller amount of parameters (9.2M vs. 12.8M) and FLOPs (1.9G vs. 2.3G).}

\begin{table}[t]
    \centering
    \caption{\gjreone{Ablation study of the proposed modules for PSLT on the ImageNet without pretraining. ``V1'' and ``V2'' denote the top-1 accuracy on ImageNet and ImageNet-V2 validation sets respectively.}}
    \setlength{\tabcolsep}{0.5mm}{\begin{tabular}{c|c|c|c|c|c|c|c}
        \hline
         Module & LSA & LFFN & PAFM & \tabincell{c}{\#Params} & \#FLOPs & V1 & V2 \\
        \hline
        ViT-B \cite{dosovitskiy2020image} & \xmark & \xmark & \xmark & 86M & 55.5G & 77.9\% & 67.5\% \\
        \hline
        LSA & \cmark & \xmark & \xmark & 9.2M & 2.0G & 77.2\% & 64.8\% \\
        +LFFN & \cmark & \cmark & \xmark & 5.6M & 1.5G & 77.3\% & 64.6\% \\
        +PAFM & \cmark & \cmark & \cmark & 9.2M & 1.9G & 79.9\% & 68.6\%\\
        \hline
    \end{tabular}}
    \label{tab:ablation-modules}
\end{table}

\vspace{0.1cm}

\noindent \textbf{- Pixel-Adaptive Fusion Module.} As described in Section \ref{sec:PAFM}, to effectively integrate the features of each branch in the ladder self-attention block, PSLT adopts a pixel-adaptive fusion module. In this module, the weight of each pixel in the feature map is first multiplied by the feature; then, the weighted features are integrated in a fully connected layer to fuse the features along the channel dimension. \gjten{``AW'' and ``FC'' denote the module without one of the two steps, respectively. The experimental results in Table \ref{tab:ablation-PAFM} demonstrate the effectiveness of our proposed pixel-adaptive fusion module.}

\gjsix{Furthermore, we implemented another module (``SE'' in Table \ref{tab:ablation-PAFM}) for comparison, which only computes the weights along the channel dimension (a pointwise following a squeeze-and-excitation layer). The model with only channel adaptive weights achieves top-1 accuracy of 79.0\% with 9.2M parameters and 1.7G FLOPs.} The experimental results demonstrate that adaptive weights in both the spatial and channel dimensions significantly improves performance.

\noindent \gjreone{\textbf{- Stacking the proposed modules.} As described in Section \ref{sec:architecture}, our proposed PSLT is composed of the proposed ladder self-attention block, Light FFN and the pixel-adaptive fusion module. Table \ref{tab:ablation-modules} shows the performance of models by stacking the proposed modules one by one, which demonstrates the effectiveness of the modules. Compared to the ViT, although applying our ladder self-attetion (LSA) and Light FFN (LFFN) does not improve the performance, the number of parameters and FLOPs are largely reduced, where the parameters have been reduced by 9 times and 15 times respectively, and the FLOPs have been reduced by 27 times and 37 times respectively. The pixel-adaptive fusion module (PAFM) improves the model performance by integrating information of all branches in the block.} \gjrelast{The top-1 accuracy of our PSLT and DeiT-S on both the ImageNet and ImageNet-V2 validation set conforms to the linear fit in [53]. Our PSLT (including DeiT-S) achieves smaller improvement over ViT-B on ImageNet-V2 validation  (2\% and 1\% top-1 accuracy improvement on ImageNet and ImageNet-V2 validation set, respectively). Such a phenomenon has been similarly observed in \cite{graham2021levit, touvron2021training}. As described in \cite{shankar2020evaluating}, this shows the model generalization ability is challenged on different validation set.}


\section{Conclusion}
In this work, \gjele{we propose a ladder self-attention block with multiple branches requiring a relatively small number of parameters and FLOPs. To improve the computation efficiency and enlarge the receptive field of the ladder self-attention block, the input feature map is split into several parts along the channel dimension, and a progressive shift mechanism is proposed for long-range dependency by modelling diverse local self-attention on each branch and interacting among these branches.}  A pixel-adaptive fusion module is finally \gjele{designed} to integrate features from multiple branches with adaptive weights along both the spatial and channel dimensions. Furthermore, the ladder self-attention block adopts a Light FFN layer to decrease the number of trainable parameters and float-point operations. Based on the above designs, we develop a general-purpose backbone (PSLT) with a relatively small number of parameters and FLOPs. Overall, PSLT applies convolutional blocks in the early stages and ladder self-attention blocks in the latter stages.
PSLT achieves comparable performance with existing methods on several computer vision tasks, including image classification, object detection and person re-identification. 

\gjele{Our work explores a new perspective to model long-range interaction by effectively utilizing diverse local self-attentions, while there are recent works \cite{pan2022edgevits, chen2021mobile, maaz2022edgenext, lee2021mpvit} implement effective methods to combine the convolution and global self-attention. We find that modelling local self-attention is more environmental-friendly than modelling global self-attention.}

For future work, \gjreone{we will investigate the direct computing optimization of our PSLT for fast inference.} And we will investigate a new computation manner for self-attention because the similarity computation is time consuming. For example, we will investigate whether simple addition is powerful enough to model interactions in the self-attention computation. PSLT shows comparable performance without elaborately selecting the hyperparameters, such as the number of blocks, the channel dimension or the number of heads in the ladder self-attention blocks. Furthermore, we can automatically select parameters that significantly improve model performance with neural architecture search techniques.

\section*{Acknowledgment}
This work was supported partially by the NSFC (U21A20471, U1911401, U1811461), Guangdong NSF Project (No. 2023B1515040025, 2020B1515120085). The corresponding author and principal investigator for this paper is Wei-Shi Zheng.


%





\ifCLASSOPTIONcaptionsoff
  \newpage
\fi



\bibliographystyle{IEEEtran}
\bibliography{egbib}

\begin{thebibliography}{10}
\providecommand{\url}[1]{#1}
\csname url@samestyle\endcsname
\providecommand{\newblock}{\relax}
\providecommand{\bibinfo}[2]{#2}
\providecommand{\BIBentrySTDinterwordspacing}{\spaceskip=0pt\relax}
\providecommand{\BIBentryALTinterwordstretchfactor}{4}
\providecommand{\BIBentryALTinterwordspacing}{\spaceskip=\fontdimen2\font plus
\BIBentryALTinterwordstretchfactor\fontdimen3\font minus
  \fontdimen4\font\relax}
\providecommand{\BIBforeignlanguage}[2]{{%
\expandafter\ifx\csname l@#1\endcsname\relax
\typeout{** WARNING: IEEEtran.bst: No hyphenation pattern has been}%
\typeout{** loaded for the language `#1'. Using the pattern for}%
\typeout{** the default language instead.}%
\else
\language=\csname l@#1\endcsname
\fi
#2}}
\providecommand{\BIBdecl}{\relax}
\BIBdecl

\bibitem{krizhevsky2012imagenet}
A.~Krizhevsky, I.~Sutskever, and G.~E. Hinton, ``Imagenet classification with
  deep convolutional neural networks,'' \emph{Advances in neural information
  processing systems}, vol.~25, 2012.

\bibitem{Russakovsky2015ImageNet}
O.~Russakovsky, J.~Deng, H.~Su, J.~Krause, S.~Satheesh, S.~Ma, Z.~Huang,
  A.~Karpathy, A.~Khosla, and M.~Bernstein, ``Imagenet large scale visual
  recognition challenge,'' \emph{International Journal of Computer Vision},
  vol. 115, no.~3, pp. 211--252, 2015.

\bibitem{he2016deep}
K.~{He}, X.~{Zhang}, S.~{Ren}, and J.~{Sun}, ``Deep residual learning for image
  recognition,'' in \emph{2016 IEEE Conference on Computer Vision and Pattern
  Recognition (CVPR)}, 2016, pp. 770--778.

\bibitem{gao2019res2net}
S.-H. Gao, M.-M. Cheng, K.~Zhao, X.-Y. Zhang, M.-H. Yang, and P.~Torr,
  ``Res2net: A new multi-scale backbone architecture,'' \emph{IEEE transactions
  on pattern analysis and machine intelligence}, vol.~43, no.~2, pp. 652--662,
  2019.

\bibitem{xie2017aggregated}
S.~Xie, R.~Girshick, P.~Doll{\'a}r, Z.~Tu, and K.~He, ``Aggregated residual
  transformations for deep neural networks,'' in \emph{Proceedings of the IEEE
  conference on computer vision and pattern recognition}, 2017, pp. 1492--1500.

\bibitem{simonyan2014very}
K.~Simonyan and A.~Zisserman, ``Very deep convolutional networks for
  large-scale image recognition,'' \emph{arXiv preprint arXiv:1409.1556}, 2014.

\bibitem{huang2017densely}
G.~Huang, Z.~Liu, L.~Van Der~Maaten, and K.~Q. Weinberger, ``Densely connected
  convolutional networks,'' in \emph{Proceedings of the IEEE conference on
  computer vision and pattern recognition}, 2017, pp. 4700--4708.

\bibitem{zagoruyko2016wide}
S.~Zagoruyko and N.~Komodakis, ``Wide residual networks,'' \emph{arXiv preprint
  arXiv:1605.07146}, 2016.

\bibitem{dai2021coatnet}
Z.~Dai, H.~Liu, Q.~Le, and M.~Tan, ``Coatnet: Marrying convolution and
  attention for all data sizes,'' \emph{Advances in Neural Information
  Processing Systems}, vol.~34, 2021.

\bibitem{xiao2021early}
T.~Xiao, M.~Singh, E.~Mintun, T.~Darrell, P.~Doll{\'a}r, and R.~Girshick,
  ``Early convolutions help transformers see better,'' \emph{Advances in Neural
  Information Processing Systems}, vol.~34, pp. 30\,392--30\,400, 2021.

\bibitem{vaswani2017attention}
A.~Vaswani, N.~Shazeer, N.~Parmar, J.~Uszkoreit, L.~Jones, A.~N. Gomez,
  {\L}.~Kaiser, and I.~Polosukhin, ``Attention is all you need,''
  \emph{Advances in neural information processing systems}, vol.~30, 2017.

\bibitem{devlin2018bert}
J.~Devlin, M.-W. Chang, K.~Lee, and K.~Toutanova, ``Bert: Pre-training of deep
  bidirectional transformers for language understanding,'' \emph{arXiv preprint
  arXiv:1810.04805}, 2018.

\bibitem{guo2021cmt}
J.~Guo, K.~Han, H.~Wu, C.~Xu, Y.~Tang, C.~Xu, and Y.~Wang, ``Cmt: Convolutional
  neural networks meet vision transformers,'' \emph{arXiv preprint
  arXiv:2107.06263}, 2021.

\bibitem{wang2021pyramid}
W.~Wang, E.~Xie, X.~Li, D.-P. Fan, K.~Song, D.~Liang, T.~Lu, P.~Luo, and
  L.~Shao, ``Pyramid vision transformer: A versatile backbone for dense
  prediction without convolutions,'' in \emph{Proceedings of the IEEE/CVF
  International Conference on Computer Vision}, 2021, pp. 568--578.

\bibitem{liu2021swin}
Z.~Liu, Y.~Lin, Y.~Cao, H.~Hu, Y.~Wei, Z.~Zhang, S.~Lin, and B.~Guo, ``Swin
  transformer: Hierarchical vision transformer using shifted windows,'' in
  \emph{Proceedings of the IEEE/CVF International Conference on Computer
  Vision}, 2021, pp. 10\,012--10\,022.

\bibitem{wang2021crossformer}
W.~Wang, L.~Yao, L.~Chen, D.~Cai, X.~He, and W.~Liu, ``Crossformer: A versatile
  vision transformer based on cross-scale attention,'' \emph{arXiv e-prints},
  pp. arXiv--2108, 2021.

\bibitem{pan2022edgevits}
J.~Pan, A.~Bulat, F.~Tan, X.~Zhu, L.~Dudziak, H.~Li, G.~Tzimiropoulos, and
  B.~Martinez, ``Edgevits: Competing light-weight cnns on mobile devices with
  vision transformers,'' \emph{arXiv preprint arXiv:2205.03436}, 2022.

\bibitem{maaz2022edgenext}
M.~Maaz, A.~Shaker, H.~Cholakkal, S.~Khan, S.~W. Zamir, R.~M. Anwer, and F.~S.
  Khan, ``Edgenext: Efficiently amalgamated cnn-transformer architecture for
  mobile vision applications,'' \emph{arXiv preprint arXiv:2206.10589}, 2022.

\bibitem{mehta2021mobilevit}
S.~Mehta and M.~Rastegari, ``Mobilevit: light-weight, general-purpose, and
  mobile-friendly vision transformer,'' \emph{arXiv preprint arXiv:2110.02178},
  2021.

\bibitem{howard2017mobilenets}
A.~G. {Howard}, M.~{Zhu}, B.~{Chen}, D.~{Kalenichenko}, W.~{Wang}, T.~{Weyand},
  M.~{Andreetto}, and H.~{Adam}, ``Mobilenets: Efficient convolutional neural
  networks for mobile vision applications,'' \emph{arXiv preprint
  arXiv:1704.04861}, 2017.

\bibitem{sandler2018mobilenetv2}
M.~{Sandler}, A.~{Howard}, M.~{Zhu}, A.~{Zhmoginov}, and L.-C. {Chen},
  ``Mobilenetv2: Inverted residuals and linear bottlenecks,'' in \emph{2018
  IEEE/CVF Conference on Computer Vision and Pattern Recognition}, 2018, pp.
  4510--4520.

\bibitem{howard2019searching}
A.~Howard, M.~Sandler, G.~Chu, L.-C. Chen, B.~Chen, M.~Tan, W.~Wang, Y.~Zhu,
  R.~Pang, V.~Vasudevan \emph{et~al.}, ``Searching for mobilenetv3,'' in
  \emph{Proceedings of the IEEE International Conference on Computer Vision},
  2019, pp. 1314--1324.

\bibitem{zhang2018shufflenet}
X.~{Zhang}, X.~{Zhou}, M.~{Lin}, and J.~{Sun}, ``Shufflenet: An extremely
  efficient convolutional neural network for mobile devices,'' in \emph{2018
  IEEE/CVF Conference on Computer Vision and Pattern Recognition}, 2018, pp.
  6848--6856.

\bibitem{zoph2017neural}
B.~{Zoph} and Q.~{Le}, ``Neural architecture search with reinforcement
  learning,'' in \emph{ICLR 2017 : International Conference on Learning
  Representations 2017}, 2017.

\bibitem{guo2020single}
Z.~Guo, X.~Zhang, H.~Mu, W.~Heng, Z.~Liu, Y.~Wei, and J.~Sun, ``Single path
  one-shot neural architecture search with uniform sampling,'' in
  \emph{European Conference on Computer Vision}.\hskip 1em plus 0.5em minus
  0.4em\relax Springer, 2020, pp. 544--560.

\bibitem{li2021bossnas}
C.~Li, T.~Tang, G.~Wang, J.~Peng, B.~Wang, X.~Liang, and X.~Chang, ``Bossnas:
  Exploring hybrid cnn-transformers with block-wisely self-supervised neural
  architecture search,'' in \emph{Proceedings of the IEEE/CVF International
  Conference on Computer Vision}, 2021, pp. 12\,281--12\,291.

\bibitem{tan2019efficientnet}
M.~{Tan} and Q.~V. {Le}, ``Efficientnet: Rethinking model scaling for
  convolutional neural networks,'' in \emph{International Conference on Machine
  Learning}, 2019, pp. 6105--6114.

\bibitem{liu2019darts}
H.~{Liu}, K.~{Simonyan}, and Y.~{Yang}, ``Darts: Differentiable architecture
  search,'' in \emph{ICLR 2019 : 7th International Conference on Learning
  Representations}, 2019.

\bibitem{li2021involution}
D.~Li, J.~Hu, C.~Wang, X.~Li, Q.~She, L.~Zhu, T.~Zhang, and Q.~Chen,
  ``Involution: Inverting the inherence of convolution for visual
  recognition,'' in \emph{Proceedings of the IEEE/CVF Conference on Computer
  Vision and Pattern Recognition}, 2021, pp. 12\,321--12\,330.

\bibitem{dai2017deformable}
J.~Dai, H.~Qi, Y.~Xiong, Y.~Li, G.~Zhang, H.~Hu, and Y.~Wei, ``Deformable
  convolutional networks,'' in \emph{Proceedings of the IEEE international
  conference on computer vision}, 2017, pp. 764--773.

\bibitem{dosovitskiy2020image}
A.~Dosovitskiy, L.~Beyer, A.~Kolesnikov, D.~Weissenborn, X.~Zhai,
  T.~Unterthiner, M.~Dehghani, M.~Minderer, G.~Heigold, S.~Gelly \emph{et~al.},
  ``An image is worth 16x16 words: Transformers for image recognition at
  scale,'' \emph{arXiv preprint arXiv:2010.11929}, 2020.

\bibitem{yue2021vision}
X.~Yue, S.~Sun, Z.~Kuang, M.~Wei, P.~H. Torr, W.~Zhang, and D.~Lin, ``Vision
  transformer with progressive sampling,'' in \emph{Proceedings of the IEEE/CVF
  International Conference on Computer Vision}, 2021, pp. 387--396.

\bibitem{yuan2021volo}
L.~Yuan, Q.~Hou, Z.~Jiang, J.~Feng, and S.~Yan, ``Volo: Vision outlooker for
  visual recognition,'' \emph{arXiv preprint arXiv:2106.13112}, 2021.

\bibitem{li2022efficientformer}
Y.~Li, G.~Yuan, Y.~Wen, E.~Hu, G.~Evangelidis, S.~Tulyakov, Y.~Wang, and
  J.~Ren, ``Efficientformer: Vision transformers at mobilenet speed,''
  \emph{arXiv preprint arXiv:2206.01191}, 2022.

\bibitem{wu2021cvt}
H.~Wu, B.~Xiao, N.~Codella, M.~Liu, X.~Dai, L.~Yuan, and L.~Zhang, ``Cvt:
  Introducing convolutions to vision transformers,'' in \emph{Proceedings of
  the IEEE/CVF International Conference on Computer Vision}, 2021, pp. 22--31.

\bibitem{peng2021conformer}
Z.~Peng, W.~Huang, S.~Gu, L.~Xie, Y.~Wang, J.~Jiao, and Q.~Ye, ``Conformer:
  Local features coupling global representations for visual recognition,'' in
  \emph{Proceedings of the IEEE/CVF International Conference on Computer
  Vision}, 2021, pp. 367--376.

\bibitem{graham2021levit}
B.~Graham, A.~El-Nouby, H.~Touvron, P.~Stock, A.~Joulin, H.~J{\'e}gou, and
  M.~Douze, ``Levit: a vision transformer in convnet's clothing for faster
  inference,'' in \emph{Proceedings of the IEEE/CVF International Conference on
  Computer Vision}, 2021, pp. 12\,259--12\,269.

\bibitem{chen2021mobile}
Y.~Chen, X.~Dai, D.~Chen, M.~Liu, X.~Dong, L.~Yuan, and Z.~Liu,
  ``Mobile-former: Bridging mobilenet and transformer,'' \emph{arXiv preprint
  arXiv:2108.05895}, 2021.

\bibitem{heo2021rethinking}
B.~Heo, S.~Yun, D.~Han, S.~Chun, J.~Choe, and S.~J. Oh, ``Rethinking spatial
  dimensions of vision transformers,'' in \emph{Proceedings of the IEEE/CVF
  International Conference on Computer Vision}, 2021, pp. 11\,936--11\,945.

\bibitem{chen2021crossvit}
C.-F.~R. Chen, Q.~Fan, and R.~Panda, ``Crossvit: Cross-attention multi-scale
  vision transformer for image classification,'' in \emph{Proceedings of the
  IEEE/CVF international conference on computer vision}, 2021, pp. 357--366.

\bibitem{fan2021multiscale}
H.~Fan, B.~Xiong, K.~Mangalam, Y.~Li, Z.~Yan, J.~Malik, and C.~Feichtenhofer,
  ``Multiscale vision transformers,'' in \emph{Proceedings of the IEEE/CVF
  International Conference on Computer Vision}, 2021, pp. 6824--6835.

\bibitem{yuan2021tokens}
L.~Yuan, Y.~Chen, T.~Wang, W.~Yu, Y.~Shi, Z.-H. Jiang, F.~E. Tay, J.~Feng, and
  S.~Yan, ``Tokens-to-token vit: Training vision transformers from scratch on
  imagenet,'' in \emph{Proceedings of the IEEE/CVF International Conference on
  Computer Vision}, 2021, pp. 558--567.

\bibitem{yu2021glance}
Q.~Yu, Y.~Xia, Y.~Bai, Y.~Lu, A.~L. Yuille, and W.~Shen, ``Glance-and-gaze
  vision transformer,'' \emph{Advances in Neural Information Processing
  Systems}, vol.~34, 2021.

\bibitem{xia2022vision}
Z.~Xia, X.~Pan, S.~Song, L.~E. Li, and G.~Huang, ``Vision transformer with
  deformable attention,'' in \emph{Proceedings of the IEEE/CVF Conference on
  Computer Vision and Pattern Recognition}, 2022, pp. 4794--4803.

\bibitem{chu2021twins}
X.~Chu, Z.~Tian, Y.~Wang, B.~Zhang, H.~Ren, X.~Wei, H.~Xia, and C.~Shen,
  ``Twins: Revisiting the design of spatial attention in vision transformers,''
  \emph{Advances in Neural Information Processing Systems}, vol.~34, 2021.

\bibitem{zhang2021rest}
Q.~Zhang and Y.-B. Yang, ``Rest: An efficient transformer for visual
  recognition,'' \emph{Advances in Neural Information Processing Systems},
  vol.~34, pp. 15\,475--15\,485, 2021.

\bibitem{dong2022cswin}
X.~Dong, J.~Bao, D.~Chen, W.~Zhang, N.~Yu, L.~Yuan, D.~Chen, and B.~Guo,
  ``Cswin transformer: A general vision transformer backbone with cross-shaped
  windows,'' in \emph{Proceedings of the IEEE/CVF Conference on Computer Vision
  and Pattern Recognition}, 2022, pp. 12\,124--12\,134.

\bibitem{lee2021mpvit}
Y.~Lee, J.~Kim, J.~Willette, and S.~J. Hwang, ``Mpvit: Multi-path vision
  transformer for dense prediction,'' \emph{arXiv preprint arXiv:2112.11010},
  2021.

\bibitem{yang2022lite}
C.~Yang, Y.~Wang, J.~Zhang, H.~Zhang, Z.~Wei, Z.~Lin, and A.~Yuille, ``Lite
  vision transformer with enhanced self-attention,'' in \emph{Proceedings of
  the IEEE/CVF Conference on Computer Vision and Pattern Recognition}, 2022,
  pp. 11\,998--12\,008.

\bibitem{hu2018squeeze}
J.~{Hu}, L.~{Shen}, S.~{Albanie}, G.~{Sun}, and E.~{Wu},
  ``Squeeze-and-excitation networks,'' in \emph{2018 IEEE/CVF Conference on
  Computer Vision and Pattern Recognition}, vol.~42, no.~8, 2018, pp.
  2011--2023.

\bibitem{lin2014microsoft}
T.-Y. Lin, M.~Maire, S.~Belongie, J.~Hays, P.~Perona, D.~Ramanan,
  P.~Doll{\'a}r, and C.~L. Zitnick, ``Microsoft coco: Common objects in
  context,'' in \emph{European conference on computer vision}.\hskip 1em plus
  0.5em minus 0.4em\relax Springer, 2014, pp. 740--755.

\bibitem{zheng2015scalable}
L.~{Zheng}, L.~{Shen}, L.~{Tian}, S.~{Wang}, J.~{Wang}, and Q.~{Tian},
  ``Scalable person re-identification: A benchmark,'' in \emph{2015 IEEE
  International Conference on Computer Vision (ICCV)}, 2015, pp. 1116--1124.

\bibitem{recht2019imagenet}
B.~Recht, R.~Roelofs, L.~Schmidt, and V.~Shankar, ``Do imagenet classifiers
  generalize to imagenet?'' in \emph{International Conference on Machine
  Learning}.\hskip 1em plus 0.5em minus 0.4em\relax PMLR, 2019, pp. 5389--5400.

\bibitem{szegedy2015going}
C.~Szegedy, W.~Liu, Y.~Jia, P.~Sermanet, S.~Reed, D.~Anguelov, D.~Erhan,
  V.~Vanhoucke, and A.~Rabinovich, ``Going deeper with convolutions,'' in
  \emph{Proceedings of the IEEE conference on computer vision and pattern
  recognition}, 2015, pp. 1--9.

\bibitem{zhang2017mixup}
H.~Zhang, M.~Cisse, Y.~N. Dauphin, and D.~Lopez-Paz, ``mixup: Beyond empirical
  risk minimization,'' \emph{arXiv preprint arXiv:1710.09412}, 2017.

\bibitem{cubuk2019autoaugment}
E.~D. Cubuk, B.~Zoph, D.~Mane, V.~Vasudevan, and Q.~V. Le, ``Autoaugment:
  Learning augmentation strategies from data,'' in \emph{Proceedings of the
  IEEE/CVF Conference on Computer Vision and Pattern Recognition}, 2019, pp.
  113--123.

\bibitem{zhong2020random}
Z.~Zhong, L.~Zheng, G.~Kang, S.~Li, and Y.~Yang, ``Random erasing data
  augmentation,'' in \emph{Proceedings of the AAAI conference on artificial
  intelligence}, vol.~34, no.~07, 2020, pp. 13\,001--13\,008.

\bibitem{szegedy2016rethinking}
C.~Szegedy, V.~Vanhoucke, S.~Ioffe, J.~Shlens, and Z.~Wojna, ``Rethinking the
  inception architecture for computer vision,'' in \emph{Proceedings of the
  IEEE conference on computer vision and pattern recognition}, 2016, pp.
  2818--2826.

\bibitem{loshchilov2017decoupled}
I.~Loshchilov and F.~Hutter, ``Decoupled weight decay regularization,''
  \emph{arXiv preprint arXiv:1711.05101}, 2017.

\bibitem{yuan2021hrformer}
Y.~Yuan, R.~Fu, L.~Huang, W.~Lin, C.~Zhang, X.~Chen, and J.~Wang, ``Hrformer:
  High-resolution transformer for dense prediction,'' \emph{arXiv preprint
  arXiv:2110.09408}, 2021.

\bibitem{ma2020weightnet}
N.~Ma, X.~Zhang, J.~Huang, and J.~Sun, ``Weightnet: Revisiting the design space
  of weight networks,'' in \emph{European Conference on Computer Vision}.\hskip
  1em plus 0.5em minus 0.4em\relax Springer, 2020, pp. 776--792.

\bibitem{yan2021contnet}
H.~Yan, Z.~Li, W.~Li, C.~Wang, M.~Wu, and C.~Zhang, ``Contnet: Why not use
  convolution and transformer at the same time?'' \emph{arXiv preprint
  arXiv:2104.13497}, 2021.

\bibitem{xu2021co}
W.~Xu, Y.~Xu, T.~Chang, and Z.~Tu, ``Co-scale conv-attentional image
  transformers,'' in \emph{Proceedings of the IEEE/CVF International Conference
  on Computer Vision}, 2021, pp. 9981--9990.

\bibitem{ren2021shunted}
S.~Ren, D.~Zhou, S.~He, J.~Feng, and X.~Wang, ``Shunted self-attention via
  multi-scale token aggregation,'' \emph{arXiv preprint arXiv:2111.15193},
  2021.

\bibitem{hatamizadeh2022global}
A.~Hatamizadeh, H.~Yin, J.~Kautz, and P.~Molchanov, ``Global context vision
  transformers,'' \emph{arXiv e-prints}, pp. arXiv--2206, 2022.

\bibitem{yu2021metaformer}
W.~Yu, M.~Luo, P.~Zhou, C.~Si, Y.~Zhou, X.~Wang, J.~Feng, and S.~Yan,
  ``Metaformer is actually what you need for vision,'' \emph{arXiv preprint
  arXiv:2111.11418}, 2021.

\bibitem{sun2019deep}
K.~Sun, B.~Xiao, D.~Liu, and J.~Wang, ``Deep high-resolution representation
  learning for human pose estimation,'' in \emph{Proceedings of the IEEE/CVF
  Conference on Computer Vision and Pattern Recognition}, 2019, pp. 5693--5703.

\bibitem{yin2022vit}
H.~Yin, A.~Vahdat, J.~M. Alvarez, A.~Mallya, J.~Kautz, and P.~Molchanov,
  ``A-vit: Adaptive tokens for efficient vision transformer,'' in
  \emph{Proceedings of the IEEE/CVF Conference on Computer Vision and Pattern
  Recognition}, 2022, pp. 10\,809--10\,818.

\bibitem{touvron2021training}
H.~Touvron, M.~Cord, M.~Douze, F.~Massa, A.~Sablayrolles, and H.~J{\'e}gou,
  ``Training data-efficient image transformers \& distillation through
  attention,'' in \emph{International Conference on Machine Learning}.\hskip
  1em plus 0.5em minus 0.4em\relax PMLR, 2021, pp. 10\,347--10\,357.

\bibitem{krizhevsky2009learning}
A.~Krizhevsky, G.~Hinton \emph{et~al.}, ``Learning multiple layers of features
  from tiny images,'' \emph{Handbook of Systemic Autoimmune Diseases}, 2009.

\bibitem{patel2022aggregating}
K.~Patel, A.~M. Bur, F.~Li, and G.~Wang, ``Aggregating global features into
  local vision transformer,'' \emph{arXiv preprint arXiv:2201.12903}, 2022.

\bibitem{sun2021sparse}
P.~Sun, R.~Zhang, Y.~Jiang, T.~Kong, C.~Xu, W.~Zhan, M.~Tomizuka, L.~Li,
  Z.~Yuan, C.~Wang \emph{et~al.}, ``Sparse r-cnn: End-to-end object detection
  with learnable proposals,'' in \emph{Proceedings of the IEEE/CVF conference
  on computer vision and pattern recognition}, 2021, pp. 14\,454--14\,463.

\bibitem{lin2017focal}
T.-Y. Lin, P.~Goyal, R.~Girshick, K.~He, and P.~Doll{\'a}r, ``Focal loss for
  dense object detection,'' in \emph{Proceedings of the IEEE international
  conference on computer vision}, 2017, pp. 2980--2988.

\bibitem{he2017mask}
K.~He, G.~Gkioxari, P.~Doll{\'a}r, and R.~Girshick, ``Mask r-cnn,'' in
  \emph{Proceedings of the IEEE international conference on computer vision},
  2017, pp. 2961--2969.

\bibitem{glorot2010understanding}
X.~Glorot and Y.~Bengio, ``Understanding the difficulty of training deep
  feedforward neural networks,'' in \emph{Proceedings of the thirteenth
  international conference on artificial intelligence and statistics}.\hskip
  1em plus 0.5em minus 0.4em\relax JMLR Workshop and Conference Proceedings,
  2010, pp. 249--256.

\bibitem{zhou2017scene}
B.~Zhou, H.~Zhao, X.~Puig, S.~Fidler, A.~Barriuso, and A.~Torralba, ``Scene
  parsing through ade20k dataset,'' in \emph{Proceedings of the IEEE conference
  on computer vision and pattern recognition}, 2017, pp. 633--641.

\bibitem{kirillov2019panoptic}
A.~Kirillov, R.~Girshick, K.~He, and P.~Doll{\'a}r, ``Panoptic feature pyramid
  networks,'' in \emph{Proceedings of the IEEE/CVF Conference on Computer
  Vision and Pattern Recognition}, 2019, pp. 6399--6408.

\bibitem{yu2015multi}
F.~Yu and V.~Koltun, ``Multi-scale context aggregation by dilated
  convolutions,'' \emph{arXiv preprint arXiv:1511.07122}, 2015.

\bibitem{xiao2018unified}
T.~Xiao, Y.~Liu, B.~Zhou, Y.~Jiang, and J.~Sun, ``Unified perceptual parsing
  for scene understanding,'' in \emph{Proceedings of the European conference on
  computer vision (ECCV)}, 2018, pp. 418--434.

\bibitem{luo2019bag}
H.~{Luo}, Y.~{Gu}, X.~{Liao}, S.~{Lai}, and W.~{Jiang}, ``Bag of tricks and a
  strong baseline for deep person re-identification,'' in \emph{2019 IEEE/CVF
  Conference on Computer Vision and Pattern Recognition Workshops (CVPRW)},
  2019, pp. 0--0.

\bibitem{zhou2019omni}
K.~{Zhou}, Y.~{Yang}, A.~{Cavallaro}, and T.~{Xiang}, ``Omni-scale feature
  learning for person re-identification,'' in \emph{2019 IEEE/CVF International
  Conference on Computer Vision (ICCV)}, 2019, pp. 3702--3712.

\bibitem{li2021combined}
H.~{Li}, G.~{Wu}, and W.-S. {Zheng}, ``Combined depth space based architecture
  search for person re-identification,'' in \emph{Proceedings of the IEEE/CVF
  Conference on Computer Vision and Pattern Recognition}, 2021, pp. 6729--6738.

\bibitem{quan2019auto}
R.~{Quan}, X.~{Dong}, Y.~{Wu}, L.~{Zhu}, and Y.~{Yang}, ``Auto-reid: Searching
  for a part-aware convnet for person re-identification,'' in \emph{2019
  IEEE/CVF International Conference on Computer Vision (ICCV)}, 2019, pp.
  3749--3758.

\bibitem{he2021transreid}
S.~He, H.~Luo, P.~Wang, F.~Wang, H.~Li, and W.~Jiang, ``Transreid:
  Transformer-based object re-identification,'' in \emph{Proceedings of the
  IEEE/CVF International Conference on Computer Vision}, 2021, pp.
  15\,013--15\,022.

\bibitem{sun2018beyond}
Y.~Sun, L.~Zheng, Y.~Yang, Q.~Tian, and S.~Wang, ``Beyond part models: Person
  retrieval with refined part pooling (and a strong convolutional baseline),''
  in \emph{Proceedings of the European conference on computer vision (ECCV)},
  2018, pp. 480--496.

\bibitem{sun2019perceive}
Y.~Sun, Q.~Xu, Y.~Li, C.~Zhang, Y.~Li, S.~Wang, and J.~Sun, ``Perceive where to
  focus: Learning visibility-aware part-level features for partial person
  re-identification,'' in \emph{Proceedings of the IEEE/CVF conference on
  computer vision and pattern recognition}, 2019, pp. 393--402.

\bibitem{hermans2017in}
A.~{Hermans}, L.~{Beyer}, and B.~{Leibe}, ``In defense of the triplet loss for
  person re-identification.'' \emph{arXiv preprint arXiv:1703.07737}, 2017.

\bibitem{shankar2020evaluating}
V.~Shankar, R.~Roelofs, H.~Mania, A.~Fang, B.~Recht, and L.~Schmidt,
  ``Evaluating machine accuracy on imagenet,'' in \emph{International
  Conference on Machine Learning}.\hskip 1em plus 0.5em minus 0.4em\relax PMLR,
  2020, pp. 8634--8644.

\end{thebibliography}
%



%

\begin{IEEEbiography}[{\includegraphics[width=1in,height=1.25in,clip,keepaspectratio]{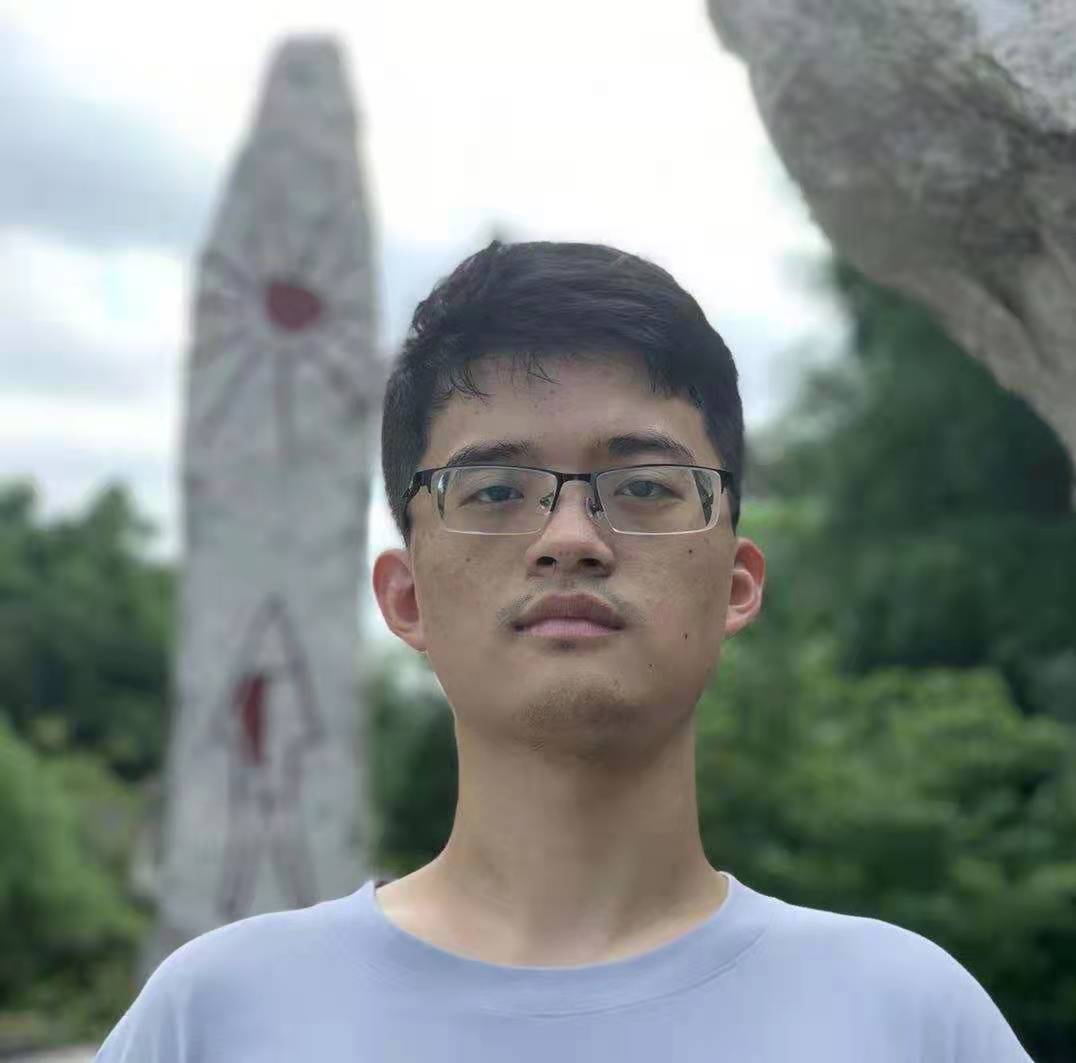}}]{Gaojie Wu} is currently a Ph.D candidate from school of computer science and engineering, Sun Yat-sen University. His research interests mainly focus on neural architecture search and computer vision. 

\end{IEEEbiography}

\begin{IEEEbiography}[{\includegraphics[width=1in,height=1.25in,clip,keepaspectratio]{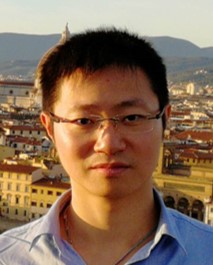}}]{Wei-Shi Zheng} is now a full Professor with Sun Yat-sen University. Dr. Zheng received his Ph.D. degree in Applied Mathematics from Sun Yat-sen University in 2008. His research interests include person/object association and activity understanding in visual surveillance, and the related large-scale machine learning algorithm. Especially, Dr. Zheng has active research on person re-identification in the last five years. He has ever joined Microsoft Research Asia Young Faculty Visiting Programme. He has ever served as area chairs of CVPR, ICCV, BMVC and IJCAI. He is an IEEE MSA TC member. He is an associate editor of the Pattern Recognition Journal. He is a recipient of the Excellent Young Scientists Fund of the National Natural Science Foundation of China, and a recipient of the Royal Society-Newton Advanced Fellowship of the United Kingdom.
\end{IEEEbiography}
\vspace{0.1cm}

\begin{IEEEbiography}[{\includegraphics[width=1in,height=1.25in,clip,keepaspectratio]{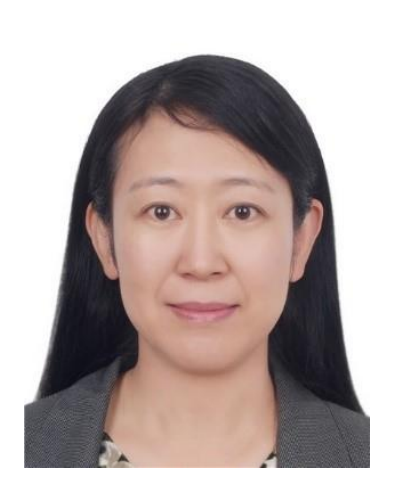}}]{Yutong Lu} is Professor in Sun Yat-sen University (SYSU), Director of National supercomputing center in Guangzhou. Her extensive research and development experience has spanned several generations of domestic supercomputers in China. Her continuing research interests include HPC, Cloud computing, storage and file system, and advanced programming environment. At present, she is devoted to the research and implementation of system and application for the convergence of HPC, Bigdata and AI on supercomputer.

\end{IEEEbiography}

\begin{IEEEbiography}[{\includegraphics[width=1in,height=1.25in,clip,keepaspectratio]{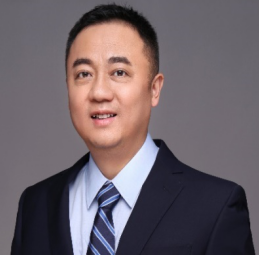}}]{Qi Tian} (Fellow, IEEE) received the B.E. degree in electronic engineering from Tsinghua University and the Ph.D. degree in electrical and computer engineering (ECE) from the University of Illinois at Urbana–Champaign (UIUC), Champaign, IL, USA. His Google citation is 50000+, with H- index 101. He was a Visiting Chaired Professor with the Center for Neural and Cognitive Computation, Tsinghua University, and a Lead Researcher with the Media Computing Group, Microsoft Research Asia (MSRA). He is currently the Chief Scientist of artificial intelligence with Huawei Cloud \& AI, a Changjiang Chaired Professor of the Ministry of Education, an Overseas Outstanding Youth, and an Overseas Expert by the Chinese Academy of Sciences.

\end{IEEEbiography}




\end{document}